%% file: main.tex
\documentclass[runningheads]{llncs}

\usepackage{eccv}

\usepackage{eccvabbrv}

\usepackage{graphicx}
\usepackage{booktabs}
\usepackage{microtype}

\usepackage[accsupp]{axessibility}  

\usepackage[pagebackref,breaklinks,colorlinks,citecolor=eccvblue]{hyperref}

\usepackage[capitalize,noabbrev]{cleveref}
\usepackage[utf8]{inputenc} 
\usepackage[T1]{fontenc}    
\usepackage{setspace}       
\usepackage{hyperref}       
\usepackage{url}            
\usepackage{booktabs}       
\usepackage{amsfonts}       
\usepackage{nicefrac}       
\usepackage{microtype}      
\usepackage{color}
\usepackage{graphicx}
\usepackage{tikz}
\usepackage{caption}
\usepackage{subcaption}
\usepackage{mathabx}
\usepackage{url}
\usepackage{algorithm}
\usepackage{algorithmic}
\usepackage{multirow}
\usepackage{xcolor,colortbl}
\usepackage{amssymb}
\usepackage{pifont}
\newcommand{\cmark}{\ding{51}}%
\newcommand{\xmark}{\ding{55}}%
\usepackage{orcidlink}

\definecolor{RowHighlight}{gray}{0.9}
\newcommand{\sname}{IDM--VTON} 

\begin{document}

\title{Improving Diffusion Models for Authentic \\ 
Virtual Try-on in the Wild}

\titlerunning{Improving Diffusion Models for Virtual Try-on}

\author{Yisol Choi\inst{1} \and
Sangkyung Kwak\inst{1} \and
Kyungmin Lee\inst{1} \and\\
Hyungwon Choi\inst{2} \and
Jinwoo Shin\inst{1}
}


\authorrunning{Y. Choi et al.}

\institute{Korea Advanced Institute of Science and Technology (KAIST) \\
\email{\{yisol.choi, skkwak9806, kyungmnlee, jinwoos\}@kaist.ac.kr} \\
\and OMNIOUS.AI\\
\email{hyungwon.choi@omnious.com}
}

\maketitle

\input{figures/main/teaser}

\begin{abstract}
    This paper considers image-based virtual try-on, which renders an image of a person wearing a curated garment, given a pair of images depicting the person and the garment, respectively. Previous works adapt existing exemplar-based inpainting diffusion models for virtual try-on to improve the naturalness of the generated visuals compared to other methods (\emph{e.g.}, GAN-based), but they fail to preserve the identity of the garments. To overcome this limitation, we propose a novel diffusion model that improves garment fidelity and generates authentic virtual try-on images. Our method, coined \emph{\sname}, uses two different modules to encode the semantics of garment image; given the base UNet of the diffusion model, 1) the high-level semantics extracted from a visual encoder are fused to the cross-attention layer, and then 2) the low-level features extracted from parallel UNet are fused to the self-attention layer. In addition, we provide detailed textual prompts for both garment and person images to enhance the authenticity of the generated visuals. Finally, we present a customization method using a pair of person-garment images, which significantly improves fidelity and authenticity. Our experimental results show that our method outperforms previous approaches (both diffusion-based and GAN-based) in preserving garment details and generating authentic virtual try-on images, both qualitatively and quantitatively. Furthermore, the proposed customization method demonstrates its effectiveness in a real-world scenario. More visualizations are available in our \href{https://idm-vton.github.io}{project page}.

\end{abstract}

\section{Introduction}\label{sec:intro}
Image-based virtual try-on (VTON) is an important computer vision task, where the goal is to render a visual of an arbitrary person dressing in a specific garment given by images. Due to its convenience and capability to provide a personalized shopping experience to E-commerce users, there is a considerable interest in synthesizing authentic virtual try-on images that exactly depict the garment given as input.
The key challenges of VTON is to fit the garment to the human body of various poses or gestures, without creating any distortions in the patterns and textures in garment~\cite{han2018viton, wang2018toward}. 

The rapid progress in generative models~\cite{goodfellow2020generative, ho2020denoising} has led the development of virtual try-on methods. Most of the primary approaches for image-based VTON are based on Generative Adversarial Networks~\cite{goodfellow2020generative, karras2019style} (GANs), where they employ a separate warping module to deform the garment to the human body and use GAN generators to blend with the target person~\cite{choi2021viton, lee2022high, men2020controllable, ge2021parser, xie2023gp, ge2021disentangled}.
However, such approaches struggle to generate high-quality images and often fail to generalize to different human images, introducing undesirable distortions to the garment. On the other hand, recently, diffusion models~\cite{ho2020denoising, song2020denoising} demonstrate superior performance in generating genuine images~\cite{podell2023sdxl, meng2021sdedit, ramesh2022hierarchical, rombach2022high} compared to GANs, showing the potential to overcome such challenges. Recent diffusion-based VTON methods~\cite{morelli2023ladi, gou2023taming, kim2023stableviton, cui2023street, li2024unihuman, ning2024picture} utilize the rich generative prior of pretrained text-to-image (T2I) diffusion models~\cite{saharia2022photorealistic, nichol2021glide, rombach2022high} to enhance the naturalness of try-on images. To identify the details of garments, they encode the semantics of garments in pseudo-words~\cite{morelli2023ladi} or use explicit warping networks~\cite{gou2023taming}. However, those methods fall short in preserving the fine-grained details of garments, \emph{e.g.}, patterns, textures, shapes, or colors, which hinders its application to real-world scenarios.

To overcome this limitation, this paper proposes an \emph{Improved Diffusion Models for Virtual Try-ON}(\sname), which significantly enhances the consistency of the garment image, while generating authentic virtual try-on images. In specific, we design a new approach in conditioning garment images to the diffusion model by designing sophisticated attention modules, which consist of two different components: 1) the image prompt adapter that encodes the high-level semantics of the garment, and 2) the UNet encoder, which we refer to as GarmentNet, that extracts low-level features to preserve fine-grained details. In addition, we propose to customize our model using a single pair of garment and person images, which further improves the visual quality of virtual try-on images, particularly in the wild scenario. 
Last but not least, we show the importance of providing detailed captions for the garment image which helps retaining the prior knowledge of T2I diffusion models.

We train our model on VITON-HD~\cite{choi2021viton} training dataset, and 
demonstrate the effectiveness of our method by showing superior results on VITON-HD and DressCode~\cite{morelli2022dress} test datasets both qualitatively and quantitatively. 
Furthermore, to simulate the real-world virtual try-on application, we have collected In-the-Wild dataset, which contains garments with intricate patterns and human images with various poses and gestures, that is largely different from the training data. Our method outperforms other methods on In-the-Wild dataset, and in particular, customizing our model notably enhances retaining the identity of the garment while generating authentic try-on images.

\section{Related Works}\label{sec:related}
\subsubsection{Image-based virtual try-on.}
Given a pair of images depicting a target person and a garment, the image-based virtual try-on aims to generate a look of the target person dressing the given garment. A line of works~\cite{choi2021viton, lee2022high, men2020controllable, ge2021parser, xie2023gp, ge2021disentangled} is based on Generative Adversarial Networks (GANs)~\cite{goodfellow2020generative}, where they first deform the garment to the person body shape, and then use the generator to put the deformed garment on the person image. While many works have attempted to reduce the mismatch between the warped garment and the person~\cite{issenhuth2020not, lee2022high, ge2021parser, choi2021viton}, those approaches lack generalization to arbitrary person images, \emph{e.g.}, with complex backgrounds or intricate poses. 

As diffusion models have shown great success~\cite{ho2020denoising, song2020score}, recent works have studied the application of diffusion models for virtual try-on. TryOnDiffusion~\cite{zhu2023tryondiffusion} proposed an architecture that uses two parallel UNets, and showed the capability of diffusion-based virtual try-on by training on the large-scale dataset. Subsequent works considered virtual try-on as an exemplar-based image inpainting problem~\cite{yang2023paint}. They fine-tuned the inpainting diffusion models on virtual try-on dataset~\cite{kim2023stableviton, morelli2023ladi, gou2023taming} to generate high-quality virtual try-on images. However, those methods still suffer from preserving meticulous details of garments, dampening their applications to real-world scenarios.

\subsubsection{Adding conditional control to diffusion models.}
While text-to-image (T2I) diffusion models~\cite{nichol2021glide, saharia2022photorealistic, rombach2022high} have shown their capability in generating high-quality images from text prompts, the inaccuracy of natural language lacks fine-grained control in image synthesis. To this end, various works have proposed to add conditional control to T2I diffusion models~\cite{zhang2023adding, ye2023ip, mou2023t2i, zhao2024uni}. 
ControlNet~\cite{zhang2023adding} and T2I-Adapter~\cite{mou2023t2i} proposed to fine-tune extra modules that encode spatial information such as edges, depth, and human pose, to control the diffusion model together with text prompts. Image prompt adapter (IP-Adapter)~\cite{ye2023ip} proposed to condition T2I diffusion models with high-level semantics of reference image to control image generation with both textual and visual prompts. 

\subsubsection{Customizing diffusion models.}
Several works have demonstrated the potential of customizing diffusion models with few personal images~\cite{ruiz2023dreambooth, gal2022image, han2023svdiff, kumari2023multi, sohn2023styledrop}. 
Together with parameter efficient fine-tuning methods~\cite{hu2021lora, houlsby2019parameter} and regularized fine-tuning objective~\cite{lee2024direct}, T2I diffusion models can be adapted to unseen examples without catastrophic forgetting.
Also, many works have studied the customization of diffusion models for various tasks such as masked image completion~\cite{tang2023realfill} or image restoration~\cite{chari2023personalized}.
In this paper, we first present a customization of diffusion models for virtual try-on, where we show it significantly improves the adaptation to real-world scenarios.

\section{Method}\label{sec:method}
\subsection{Backgrounds on Diffusion Models}\label{sec:diff}
Diffusion models~\cite{ho2020denoising, sohl2015deep} are generative models that consist of a forward process, which gradually adds Gaussian noise to the data, and a reverse process that gradually denoises a random noise to generate a sample. Suppose $\mathbf{x}_0$ be a data point (\emph{e.g.}, an image or a latent from the output of autoencoder~\cite{rombach2022high}). 
Given noise schedules $\{\alpha_t\}_{t=1}^T, \{\sigma_t\}_{t=1}^T$ for $t=1,\ldots, T$, the forward process at timestep $t$ is given by $\mathbf{x}_t = \alpha_t \mathbf{x}_0 + \sigma_t \boldsymbol{\epsilon}$, where $\boldsymbol{\epsilon}\sim\mathcal{N}(\boldsymbol{0},\mathbf{I})$ is a Gaussian noise. For sufficiently large $\sigma_T$, $\mathbf{x}_T \sim\mathcal{N}(\alpha_T\mathbf{x}_0, \sigma_T^2\mathbf{I})$ is indiscernible from pure random Gaussian noise. Then the reverse generative process initializes from $\mathbf{x}_T\sim \mathcal{N}(\boldsymbol{0},\sigma_T^2\mathbf{I})$, and sequentially denoises $\mathbf{x}_{t}$ into $\mathbf{x}_{t-1}$ for each $t$, so that $\mathbf{x}_0$ is distributed according to the data distribution.

\subsubsection{Text-to-image (T2I) diffusion models.} T2I diffusion models~\cite{saharia2022photorealistic, nichol2021glide, rombach2022high} are diffusion models that model the distribution of image conditioned on text, which is encoded by embeddings using the pretrained text encoders (\emph{e.g.}, T5~\cite{raffel2020exploring} and CLIP~\cite{radford2021learning} text encoder). While convolutional UNet~\cite{ronneberger2015u} architecture has been developed for diffusion-based generative models~\cite{ho2020denoising, song2020score}, recent works have shown the promise of fusing transformer~\cite{vaswani2017attention} architectures for UNet. Training a diffusion model is shown to be equivalent to learning a score function of a perturbed data distribution~(\emph{i.e.}, denoising score matching~\cite{hyvarinen2005estimation}), where it is often done by $\epsilon$-noise prediction loss~\cite{ho2020denoising}. Formally, given data $\mathbf{x}_0$ and text embedding $\mathbf{c}$, the training loss for the T2I diffusion model is given as follows: 
\begin{equation}
    \mathcal{L}_{\textrm{DM}}(\theta) = \mathbb{E}_{ \boldsymbol{\epsilon}\sim\mathcal{N}(\boldsymbol{0},\mathbf{I}), \,t\sim\mathcal{U}[0,T]}\big[\,\omega(t) \|\boldsymbol{\epsilon}_\theta(\mathbf{x}_t;\mathbf{c},t) - \boldsymbol{\epsilon}\|_2^2 \,\big]\text{,}
\end{equation}
where $\mathbf{x}_t = \alpha_t\mathbf{x}_0 + \sigma_t \boldsymbol{\epsilon}$ is a forward process of $\mathbf{x}_0$ and $\omega(t)$ is a weighting function at each timestep $t$. To achieve better controllability over text conditioning, it uses classifier-free guidance (CFG)~\cite{ho2022classifier}, which jointly learns unconditional and conditional. At the training stage, the text conditioning is randomly dropped out (\emph{i.e.}, giving null-text to the input), and at the inference stage, the CFG interpolates the conditional and unconditional noise output to control the strength of text conditioning:
\begin{equation}
    \boldsymbol{\hat{\epsilon}}_\theta(\mathbf{x}_t;\mathbf{c},t) = s\cdot (\boldsymbol{\epsilon}_\theta(\mathbf{x}_t;\mathbf{c},t) - \boldsymbol{\epsilon}_\theta (\mathbf{x}_t;t)) + \boldsymbol{\epsilon}_\theta(\mathbf{x}_t;t)\text{,}
\end{equation}
where $s \geq 1 $ is a guidance scale that controls the text conditioning, and $\boldsymbol{\epsilon}_\theta(\mathbf{x}_t;t)$ denotes the noise prediction with null-text embeddings. 

\input{figures/main/pipeline}

\subsubsection{Image prompt adapter~\cite{ye2023ip}.}
To condition the T2I diffusion model with a reference image, Ye et al.~\cite{ye2023ip} proposed Image Prompt Adapter (IP-Adapter), which leverages the features extracted from image encoder (\emph{e.g.}, CLIP~\cite{radford2021learning} image encoder) and attaches additional cross-attention layer to the text conditioning~\cite{rombach2022high}. 
Let us denote $Q\in\mathbb{R}^{N\times d}$ be the query matrices from the intermediate representation of UNet, and $K_c\in\mathbb{R}^{N\times d}$, $V_c\in\mathbb{R}^{N\times d}$ be the key and value matrices from the text embeddings $\mathbf{c}$, where $N$ is number of samples. 
The output of cross-attention layer is given by $\texttt{Attention}(Q,K_c,V_c) = \texttt{softmax}(\frac{QK_c^\top}{\sqrt{d}}) \cdot V_c$.
Then IP-Adapter computes key and value matrices $K_i\in\mathbb{R}^{N\times d}$, $V_i\in\mathbb{R}^{N\times d}$ from the image embeddings $\mathbf{i}$, and inserts the cross-attention layer as follows: 
\begin{equation}\label{eq:ca}
    \texttt{Attention}(Q, K_c,V_c) + \texttt{Attention}(Q, K_i, V_i)\text{.}
\end{equation}
Overall, the IP-Adapter freezes the original UNet, and fine-tunes only the (linear) projection layers of key and value matrices of image embeddings $K_i$ and $V_i$, and the linear projection layer that maps CLIP image embeddings.

\subsection{Proposed Method}\label{sec:method}
Now, we present our method for designing diffusion models for virtual try-on. Let us denote $\mathbf{x}_p$ be the image of a person, and $\mathbf{x}_g$ be the image of a garment. Our primary goal is to generate an image $\mathbf{x}_{\textrm{tr}}$ that visualizes a person from $\mathbf{x}_p$ wearing a garment in image $\mathbf{x}_g$.  It is common practice to cast virtual try-on as an exemplar-based image inpainting problem~\cite{yang2023paint}, which aims to fill the masked image with a reference image.
Here, it is important to extract relevant information of garment, and add conditional control to the diffusion model. To this end, our model is composed of three components; 
1) the base UNet (TryonNet) that processes the masked person image with its pose information, 2) the image prompt adapter (IP-Adapter) that extracts the high-level semantics of the garment, 
and 3) the garment UNet feature encoder (GarmentNet) that extracts the low-level features of the garment.
The features from GarmentNet are fused within self-attention layer of TryonNet, and then it is processed with the features of IP-Adapter through cross-attention layer. 
We provide an overview of our method in Fig.~\ref{fig:pipeline}, and provide detailed explanation of each component as follows.

\subsubsection{TryonNet.}
For the base UNet model, we consider latent diffusion model~\cite{rombach2022high}, where the diffusion generative modeling is conducted in latent space of variational autoencoder $\mathcal{E}$ and the output is passed to decoder $\mathcal{D}$ to generate an image. As of the input for our base UNet, we concatenate four components as follows: 1) the latent of person image, \emph{i.e.}, $\mathcal{E}(\mathbf{x}_p)$, 2) the (resized) mask $\mathbf{m}$ that removes the garment on the person image, 3) the latent of masked-out person image $\mathbf{x}_m = (1-\mathbf{m})\odot \mathbf{x}_p$, \emph{i.e.}, $\mathcal{E}(\mathbf{x}_m)$, following \cite{choi2021viton}, and 4) the latent of the Densepose~\cite{guler2018densepose} $\mathbf{x}_{\textrm{pose}}$ of a person image, \emph{i.e.}, $\mathcal{E}(\mathbf{x}_{\textrm{pose}})$. Then, the latents are aligned within the channel axis, where we expand the convolutional layer of UNet to 13 channels initialized with zero weights. Unlike previous works on virtual try-on with diffusion models~\cite{morelli2023ladi,kim2023stableviton, gou2023taming}, we leverage Stable Diffusion XL (SDXL)~\cite{podell2023sdxl} inpainting model~\cite{sdxlinpainting}. 

\subsubsection{Image prompt adapter.}
To condition the high-level semantics of a garment image, we leverage image prompt adapter (IP-Adapter)~\cite{ye2023ip}.  
To encode the garment image, we use frozen CLIP image encoder (\emph{i.e.}, OpenCLIP~\cite{ilharco_gabriel_2021_5143773} ViT-H/14) to extract feature and fine-tune feature projection layers and cross-attention layers, which are initialized with pretrained IP-Adapters. Note that we pass textual prompts of garments, and the cross-attention is computed as in Eq.~\eqref{eq:ca}. 

\input{figures/main/datasetcmp}
\input{figures/main/main_qual}

\subsubsection{GarmentNet.}
While we already condition the garment image using IP-Adapter, it falls short in preserving the fine-grained details of garments when it has complicated patterns or graphic prints (\emph{e.g.}, see Fig.~\ref{fig:ablation_garment}). This is because the CLIP image encoder lacks extracting the low-level features of a garment. To tackle this issue, we propose to utilize an additional UNet encoder (\emph{i.e.}, garment UNet encoder) to encode fine-details of garment images. Given the latent of a garment image $\mathcal{E}(\mathbf{x}_g)$, we pass through the (frozen) pretrained UNet encoder to obtain the intermediate representation, and concatenate with the intermediate representation from the TryonNet. Then we compute the self-attention on the concatenated features and then pass only the first-half dimensions from TryonNet. We use UNet of SDXL \cite{podell2023sdxl} for our GarmentNet, which helps utilizing the rich generative prior of the pretrained text-to-image diffusion model, and complements the low-level feature that is often ignored in the cross-attention layer of IP-Adapter. Remark that a similar approach was introduced in \cite{hu2023animate}, which used a similar attention mechanism in preserving consistency for video generation.

\subsubsection{Detailed captioning of garments.}\label{sec:caption}
While the majority of diffusion-based virtual try-on models leverage pretrained text-to-image diffusion models, they do not take the text prompts as input~\cite{gou2023taming, kim2023stableviton} or use a na\"ive text prompts~\cite{morelli2023ladi} such as ``upper garment'' for any garment image.
To fully exploit the rich generative prior of the text-to-image diffusion model, we provide a comprehensive caption~\cite{lee2024direct} that describes the detail of a garment, such as shapes or textures. 
As shown in Fig.~\ref{fig:pipeline}, we provide a comprehensive caption to the garment image (\emph{e.g.}, ``short sleeve round neck t-shirts'') and pass it to both GarmentNet (\emph{i.e.}, ``a photo of short sleeve round neck t-shirts'') and the TryonNet (\emph{i.e.}, ``model is wearing short sleeve round neck t-shirts''). 
This helps the model to encode the high-level semantics of the garment using natural language, and complements the image-based conditioning (\emph{e.g.}, see Fig.~\ref{fig:ablation_caption}). In practice, we utilize an image annotator for fashion images, and manually provide captioning with templates (see Appendix for details).

\subsubsection{Customization of \sname.}
While our model is able to capture the fine-details of garments, it often struggles when the person image $\mathbf{x}_p$ or the garment image $\mathbf{x}_g$ is different from training distribution (\emph{e.g.}, see Fig.~\ref{fig:datasetcmp}). To this end, inspired by text-to-image personalization methods~\cite{ruiz2023dreambooth, tang2023realfill, chari2023personalized}, we propose an effective customization of {\sname} by fine-tuning TryonNet with a pair of garment and person images. It is straightforward to fine-tune {\sname} when we have a pair of images $\mathbf{x}_p$ and $\mathbf{x}_g$, where the person in $\mathbf{x}_p$ is wearing the garment in $\mathbf{x}_g$. 
On the other hand, if we only have $\mathbf{x}_p$, we segment the garment and white-out backgrounds to obtain $\mathbf{x}_g$. Remark that we fine-tune only the attention layers of the up-blocks (\emph{i.e.}, decoder layers) of TryonNet, which empirically works well (\emph{e.g.}, see Table~\ref{tab:adapt}). 

\input{figures/table_vitonhd}

\section{Experiment} 
\input{figures/main/qualwildmain}

\subsection{Experimental Setup}
\subsubsection{Implementation details.}\label{sec:exp_detail}
We use SDXL inpainting model~\cite{sdxlinpainting} for our TryonNet, pretrained IP-Adapter~\cite{ye2023ip} for image adapter and UNet of SDXL~\cite{podell2023sdxl} for our GarmentNet. We train our model using VITON-HD~\cite{choi2021viton} train dataset, which contains $11,647$ person-garment image pairs at $1024\times768$ resolution. We train our model using Adam~\cite{kingma2014adam} optimizer with batch size of $24$ and learning rate of 1e-5 for $130$ epochs. For customization, we fine-tune our model using Adam optimizer with learning rate 1e-6 for $100$ steps. 

\subsubsection{Evaluation dataset.}\label{datasetexp}
We evaluate our model on public VITON-HD~\cite{choi2021viton} and DressCode~\cite{morelli2022dress} test datasets. Furthermore, we internally collected In-the-Wild dataset, which is crawled from the web to assess the generalization of model to the real-world scenario. While VITON-HD and DressCode contain person images of simple poses and solid backgrounds, our In-the-Wild dataset has person images of various poses and gestures with complex backgrounds. Also, the garments are of various patterns and logos, making virtual try-on more challenging than VITON-HD and DressCode (\emph{e.g.}, see Fig.~\ref{fig:datasetcmp}). Furthermore, In-the-Wild dataset contains multiple person images per garment, so it is possible to evaluate the results when we customize our network with each garment. Specifically, it consists of 62 images of upper garments and 312 images of people wearing those garments, where there are 4-6 images of person wearing the same garment. To conduct customization experiment, we select one person image for each garment, and use the left $250$ person images for evaluation. 

\subsubsection{Baselines.}
We compare our method with GAN-based methods (HR-VITON~\cite{lee2022high} and GP-VTON~\cite{xie2023gp}), 
and diffusion-based methods (LaDI-VTON~\cite{morelli2023ladi}, DCI-VTON~\cite{gou2023taming}, and StableVITON~\cite{kim2023stableviton}). HR-VITON and GP-VTON have a separate warping module to estimate the fitted garment on the target person, generating try-on images with GAN using the garment as input. 
LaDI-VTON, DCI-VTON, and StableVITON are most relevant baselines of ours, where they also utilize pretrained Stable Diffusion (SD). In specific, LaDI-VTON and DCI-VTON employ separate warping modules to condition the garment, while StableVITON utilizes the SD Encoder for garment conditioning. We use the pretrained checkpoints provided in their official repositories to generate images. For a fair comparison, we generate images of resolution $1024\times768$ images if available, otherwise generate images of resolution $512\times384$ and upscale to $1024\times768$ using various methods, \emph{e.g.}, interpolation or super resolution~\cite{wang2021real}, and report the best one.

\vspace{-15pt}
\subsubsection{Evaluation metrics.}
For quantitative evaluation, we measure low-level reconstruction scores such as LPIPS~\cite{zhang2018unreasonable} and SSIM~\cite{wang2004image}, and high-level image similarity using CLIP~\cite{radford2021learning} image similarity score.
To evaluate the authenticity of generated images, we compute French\'et Inception Distance (FID)~\cite{heusel2017gans} score on VITON-HD and DressCode test dataset.

\input{figures/table_wild}

\subsection{Results on Public Dataset}\label{sec:result1}
\subsubsection{Qualitative results.}
Fig.~\ref{fig:qual_main} shows the visual comparisons of {\sname} with other methods on VITON-HD and DressCode test datasets. We see that {\sname} preserves both low-level details and high-level semantics, while others struggle to capture. While GAN-based methods (\emph{i.e.}, HR-VITON and GP-VTON) show comparable performance in capturing fine details of garment, they fall shorts in generating natural human images compared to diffusion-based methods. On the other hand, prior diffusion-based methods (\emph{i.e.}, LaDI-VTON, DCI-VTON, and StableVITON) fail to preserve the details of garments (\emph{e.g.}, the off-shoulder t-shirts in first row, or the details of sweatshirts in second row of VITON-HD dataset). Also, we observe that {\sname} significantly outperforms other methods in generalization to the DressCode dataset. Especially, GAN-based methods show inferior performance on the DressCode dataset, showing its poor generalization ability. While prior diffusion-based methods show better images in generating natural images than GAN-based, the generated images show lower consistency with respect to given garment image compared to {\sname}.

\subsubsection{Quantitative results.}
Tab.~\ref{tab:main1} shows the quantitative comparisons between {\sname} (ours) and other methods on VITON-HD and DressCode test datasets. 
When compared to GAN-based methods, we observe that {\sname} shows comparable performance in reconstruction scores (LPIPS and SSIM) and outperforms in FID and CLIP image similarity score on VITON-HD test dataset.
However, the performance of GAN-based methods significantly degrades when tested on DressCode dataset, which we have observed in Fig.~\ref{fig:qual_main}. 
We see that {\sname} consistently outperforms prior diffusion-based methods in both VITON-HD and DressCode dataset. 

\subsection{Results on In-the-Wild Dataset}\label{sec:result2}
Here, we evaluate our methods on challenging In-the-Wild dataset, where we compare with other diffusion-based VTON methods. In addition, we show the results of customization using a single pair of garment-person images. 
We also test customization on StableVITON~\cite{kim2023stableviton}.

\subsubsection{Qualitative results.}
Fig.~\ref{fig:qual_wildmain} shows the qualitative results of {\sname} (ours) compared to other baselines\footnote{We do not compare with GP-VTON as it uses private parsing modules. Comparison with HR-VITON is in the appendix.}. 
We see that {\sname} generates more authentic images than LaDI-VTON, DCI-VTON, and StableVITON, yet it struggles to preserve the intricate patterns of garments. 
By customizing {\sname}, we obtain virtual try-on images with a high degree of consistency to the garment (\emph{e.g.}, the logos and the text renderings), but we find the customization does not work well for StableVITON.

\subsubsection{Quantitative results.}
Tab.~\ref{tab:table_wild} reports the quantitative comparisons between our {\sname} and other methods. We also provide comparisons between {\sname} and StableVITON using customization (\emph{i.e.}, {\sname}$^\ddag$ and StableVITON$^\ddag$). 
We see that {\sname} outperforms other methods on all metrics, even outperforming StableVITON with customization. Furthermore, we observe that {\sname} with customization significantly outperforms others, showing its generalization capability.  

\subsection{Ablative Studies}\label{sec:abl}

\input{figures/main/ablation_garment}

\input{figures/main/ablation_caption}

\subsubsection{Effect of GarmentNet.}
We conduct an ablation study on the effect of GarmentNet in preserving the fine-grained details of a garment.
For comparison, we train another model without using GarmentNet (\emph{i.e.}, using only IP-Adapter to encode garment image). 
The qualitative results are in Fig.~\ref{fig:ablation_garment}.
While training with only IP-Adapter is able to generate authentic virtual try-on images that are semantically consistent with the garment images, we see that using GarmentNet significantly improves preserving the identity of the garment (\emph{e.g.}, the graphics of t-shirts). 
In addition, we conduct a quantitative evaluation on VITON-HD test dataset by measuring LPIPS, SSIM and CLIP image similarity score. 
In Tab.~\ref{tab:ablunet}, we see that using GarmentNet quantitatively improves reconstruction scores, as well as image similarity scores, which is aligned with our qualitative results.

\subsubsection{Effect of detailed caption.} 
To validate the effectiveness of detailed captioning on garments, we train an identical model to {\sname} except that using na\"ive caption~\cite{morelli2023ladi}, and compare with {\sname}.
In specific, na\"ive caption provides an unified descriptions of all upper garments (\emph{e.g.}, ``model is wearing an upper garment''), while our detailed captions provide specific descriptions of each garment.
Fig.~\ref{fig:ablation_caption} shows generated virtual try-on images using each model trained with na\"ive caption (left) and detailed captions (right).
We observe that the model trained with detailed captions generates images with higher consistency to the garment (\emph{e.g.}, meshes around the neck and arms). Thus, the usage of detailed captions complements the potential inaccuracy of image-based virtual try-on by exploiting the generative prior of text-to-image diffusion model.

\input{figures/table_unetclip}

\subsubsection{Ablation on customization.}
During customization, we fine-tune the up-blocks of TryonNet (\emph{i.e.}, decoder layers). Here, we provide an ablation study on the choice of fine-tuning layer for customization. 
We compare default {\sname} (\emph{i.e.}, no customization), fine-tuning all UNet layers, using low rank adaptation (LoRA)~\cite{hu2021lora}, and fine-tuning decoder layers. 
In Tab.~\ref{tab:adapt}, we see that all customization methods improve reconstruction score (LPIPS, SSIM) compared to default {\sname}, while showing comparable performance in CLIP image similarity score. We see that fine-tuning decoder layers perform the best among customization methods, while fine-tuning all UNet layers shows comparable performance. While fine-tuning with LoRA is more efficient, we find it underperforms than fine-tuning decoder layers. 


\section{Conclusion}
In this paper, we present \sname, a novel design of diffusion models for authentic virtual try-on, particularly in the wild scenario. 
We incorporate two separate modules to encode the garment image, \emph{i.e.}, visual encoder and parallel UNet, which effectively encode high-level semantics and low-level features to the base UNet, respectively.
In order to improve the virtual try-on on real world scenario, we propose to customize our model by fine-tuning the decoder layers of UNet given a pair of garment-person images.
We also leverage detailed natural language descriptions for garments, which help generating authentic virtual try-on images. 
Extensive experiments on various datasets show the superiority of our method over prior works in both preserving details of garment, and generating high fidelity images. 
In particular, we demonstrate the potential of our method in virtual try-on in the wild.

\subsubsection{Potential negative impact.}
The virtual try-on technology comes with benefits and pitfalls - the tool could be helpful for users to effectively visualize their look with given garment. However, it requires the users to be responsible for protecting the ownership, and avoid malicious use.

\subsubsection{Limitation.}
\sname~struggles to preserve human attributes on masked regions such as tattoos or skin moles. One can design a method for conditioning those human attributes of when generating try-on images. Also, while we exploit the prior knowledge of T2I diffusion models by providing comprehensive captions, exploring broader applications to fully utilize the prior, such as controlling garment generation through textual prompts, is left as future work.

\section*{Acknowledgement}
This work was supported by Institute for Information \& communications Technology Promotion (IITP) grant funded by the Korea government (MSIT) (No.2019-0-00075 Artificial Intelligence Graduate School Program(KAIST); No.RS-2021-II212068, Artificial Intelligence Innovation Hub). 



%
%
\newpage
\bibliographystyle{splncs04}
\bibliography{main}

\input{appendix}

\end{document}

%% file: figures/main/teaser.tex
\begin{figure}[ht]
    \small\centering
    \includegraphics[width=0.98\textwidth]{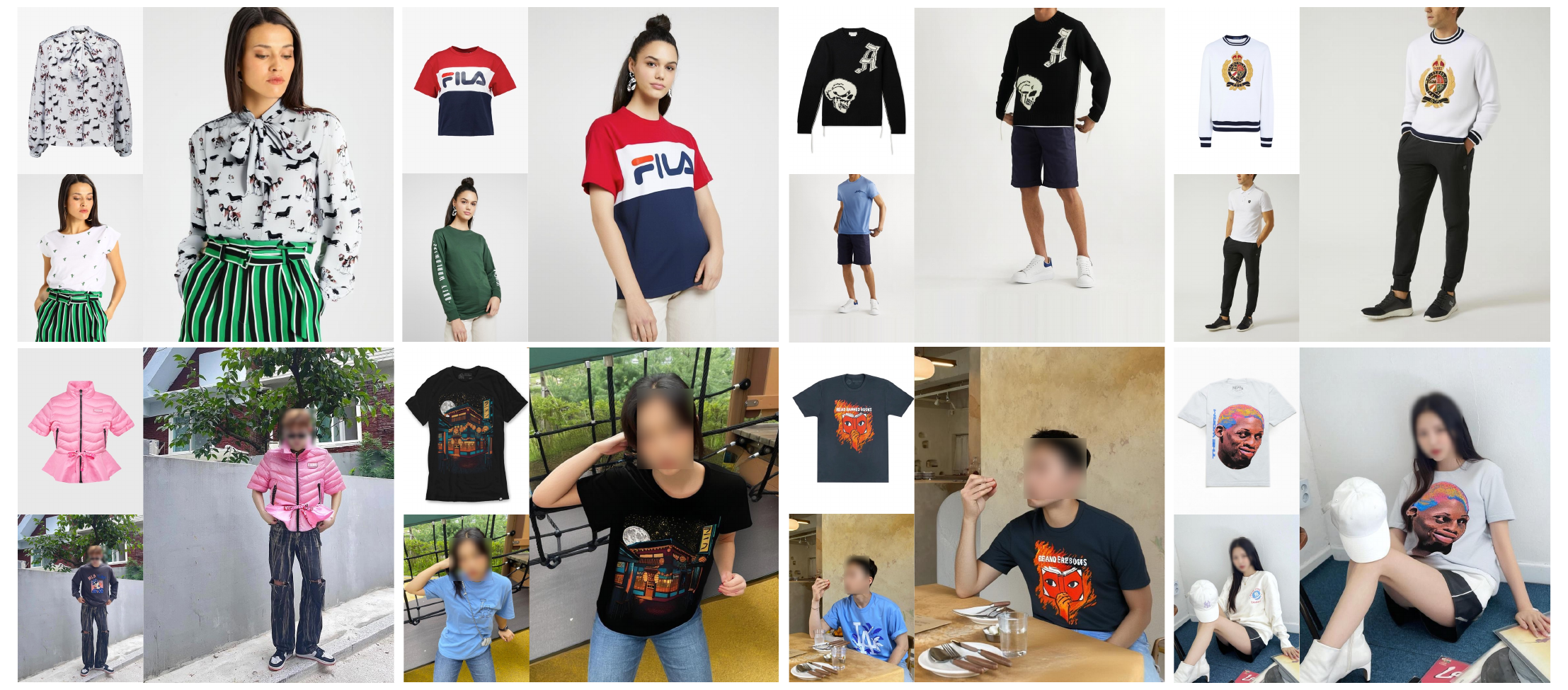}
    \vspace{-10pt}
    \caption{
    Virtual try-on images generated by using our {\sname} on VITON-HD~\cite{choi2021viton} (top row, first and second column), DressCode~\cite{morelli2022dress} (top row, third and fourth column), and collected In-the-Wild dataset (bottom row).     
    Best viewed on a zoomed, color monitor. 
    }
    \label{fig:teaser}
\end{figure}
\vspace{-10pt}

%% file: figures/main/pipeline.tex
\begin{figure}[t]
    \small\centering
    \includegraphics[width=0.98\textwidth]{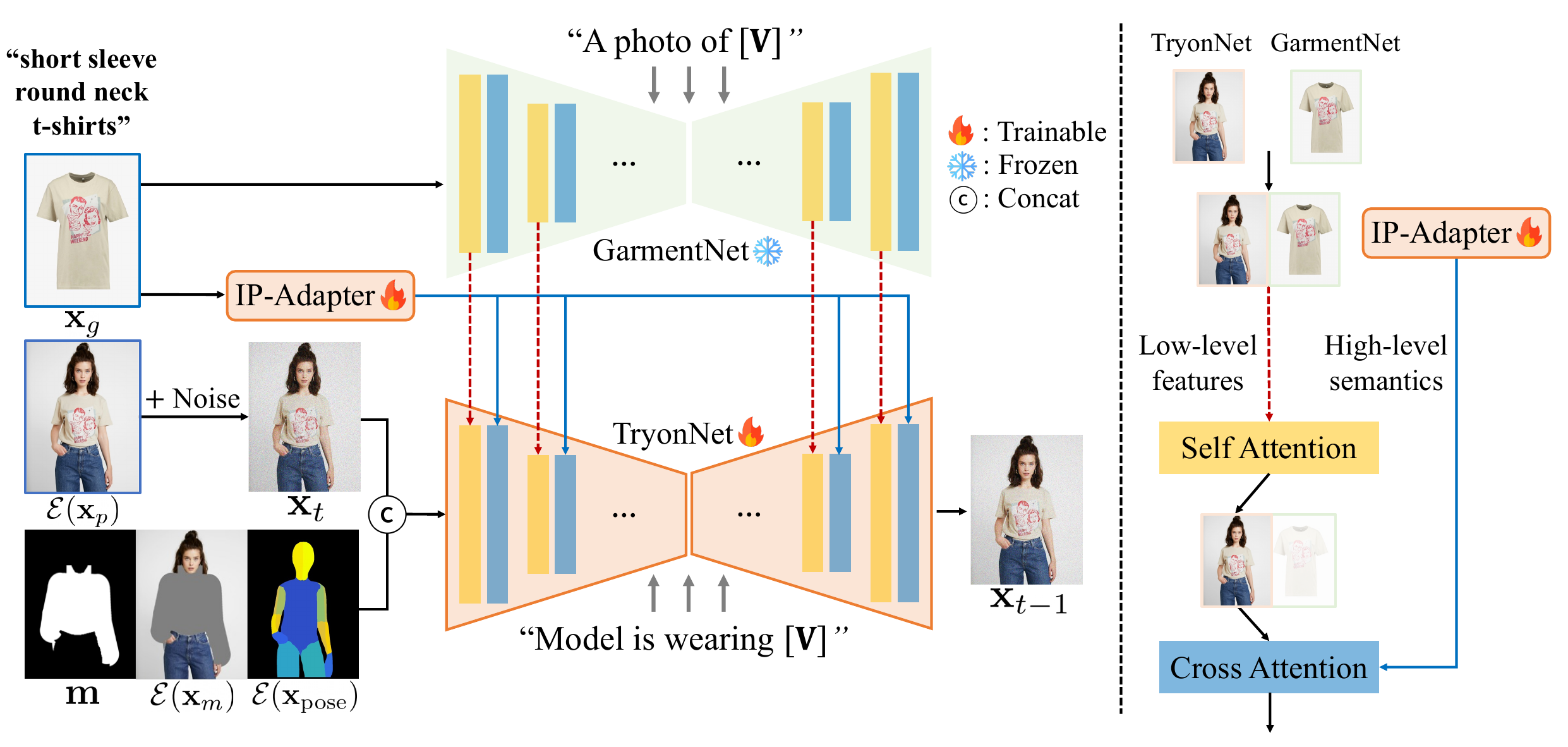}
    \vspace{-5pt}
    \caption{
    \textbf{Overview of \sname.}
    We demonstrate the proposed model architecture and details on the attention modules.
    \textbf{(Left)} Our model consists of 1) TryonNet which is a main UNet that processes person image, 2) image prompt adapter (IP-Adapter)~\cite{ye2023ip} that encodes high-level semantics of garment image $\mathbf{x}_g$, and 3) GarmentNet that encodes low-level features of $\mathbf{x}_g$. As of input for UNet, we concatenate the noised latents $\mathbf{x}_t$ of latents $\mathcal{E}(\mathbf{x}_p)$ with the segmentation mask $\mathbf{m}$, masked image $\mathcal{E}(\mathbf{x}_m)$, and Densepose~\cite{guler2018densepose} $\mathcal{E}(\mathbf{x}_{\textrm{pose}})$. We provide a detailed caption to the garment (\emph{e.g.}, [V]: ``short sleeve round neck t-shirts''). Then it is used for input prompt of GarmentNet (\emph{e.g.}, ``A photo of [V]'') and TryonNet (\emph{e.g.}, ``Model is wearing [V]''). \textbf{(Right)} The intermediate features of TryonNet and GarmentNet are concatenated and passed to the self-attention layer, and we use the first half (\emph{i.e.}, that from TryonNet) of the output. Then we fuse the output with features from text encoder and IP-Adapter by cross-attention layer. We fine-tune the TryonNet and the IP-Adapter modules, and freeze other components. 
    } 
    \label{fig:pipeline}
    \vspace{-10pt}
\end{figure}

%% file: figures/main/datasetcmp.tex
\begin{figure}[t]
    \small\centering
    \includegraphics[width=0.99\linewidth]{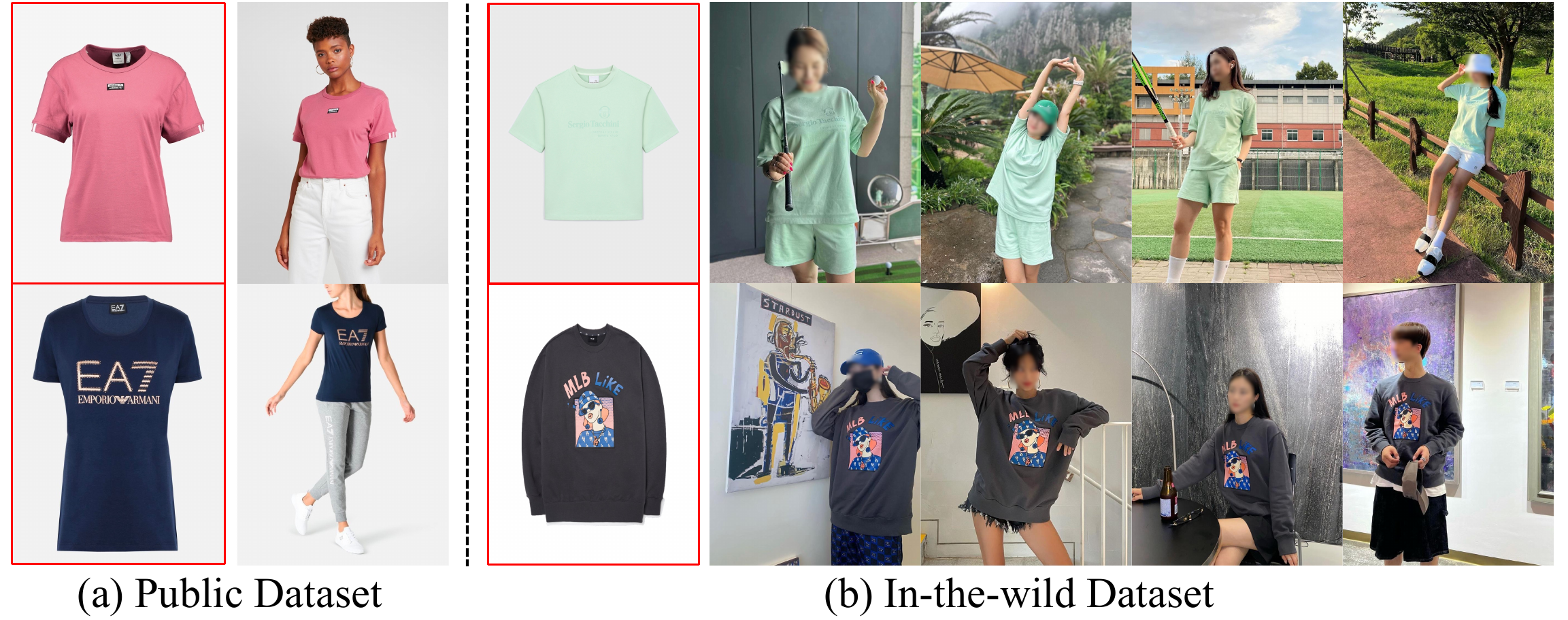}
    \vspace{-5pt}
    \caption{
    \textbf{Comparisons between datasets used in our experiments.} For evaluation, we test on (a) public dataset, including VITON-HD~\cite{choi2021viton} and DressCode~\cite{morelli2022dress}, and (b) In-the-Wild dataset, which we internally collected from real E-commerce setup. We remark that the In-the-Wild dataset contains more intricate patterns and logos in garment image, and diverse backgrounds, and poses in person image. 
    }
    \label{fig:datasetcmp}
    \vspace{-10pt}
\end{figure}

%% file: figures/main/main_qual.tex
\begin{figure}[t]
    \small\centering
    \includegraphics[width=0.95\linewidth]{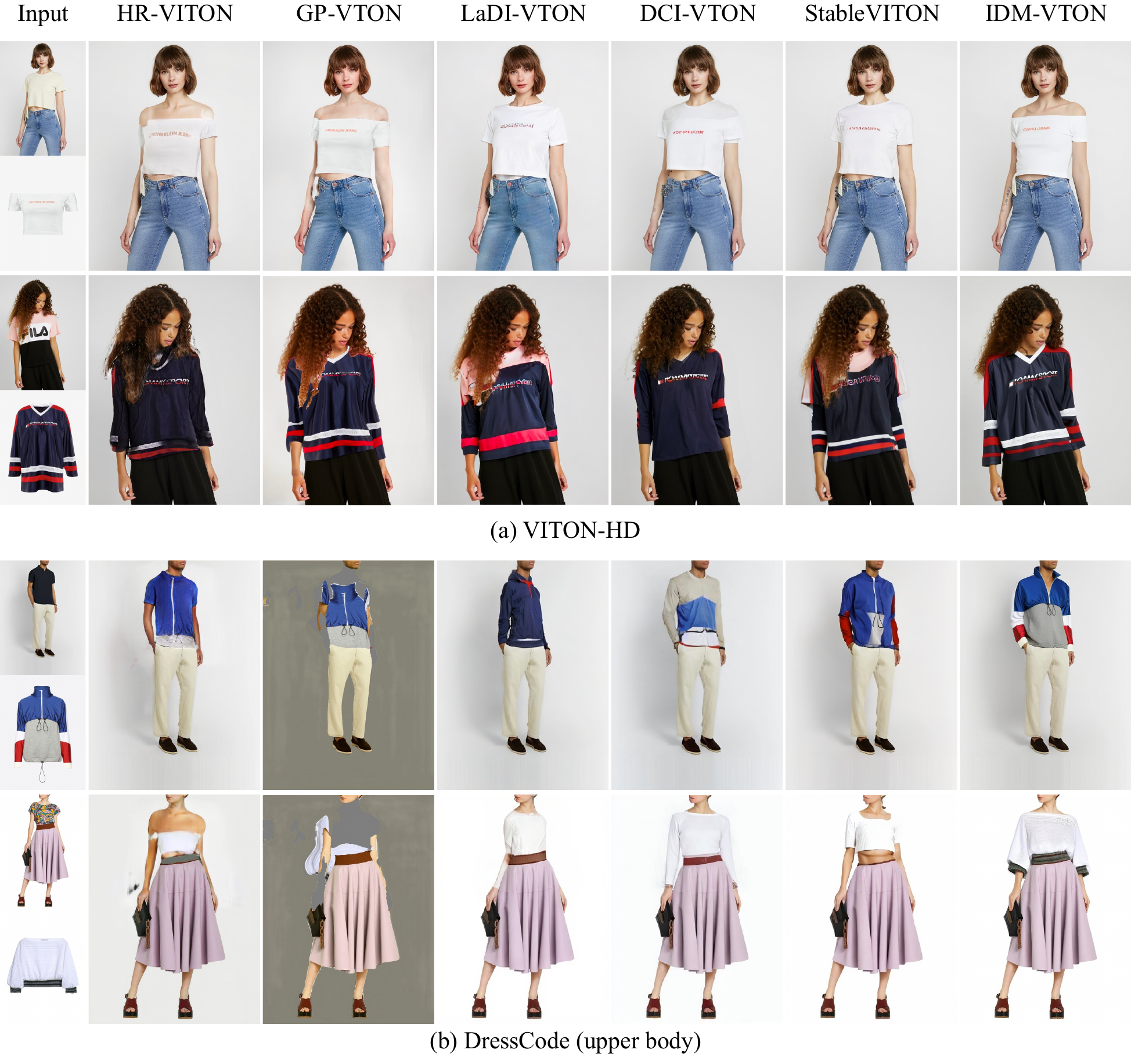}
    \vspace{-5pt}
    \caption{
    \textbf{Qualitative results on VITON-HD and DressCode dataset.} 
    We show generated virtual try-on images using {\sname} (ours) compared with other methods on (a) VITON-HD~\cite{choi2021viton}, and (b) DressCode (upper body)~\cite{morelli2022dress} test datasets. We see that {\sname} outperforms others in generating authentic images and preserving fine-grained details of garment. Best viewed in zoomed, color monitor.
    }
    \label{fig:qual_main}
    \vspace{-10pt}
\end{figure}

%% file: figures/table_vitonhd.tex
\begin{table}[t]
    \centering\small
    \caption{\textbf{Quantitative results on VITON-HD dataset.}
    We evaluate the reconstruction scores (\emph{e.g.}, LPIPS, SSIM) for low-level similarity, CLIP image similarity score (CLIP-I) for high-level semantic similarity, and FID score for image fidelity. We compare with GAN-based virtual try-on methods (\emph{e.g.}, HR-VTON and GP-VTON), and Diffusion-based virtual try-on methods (\emph{e.g.}, LaDI-VTON, DCI-VTON, StableVITON).
    \textbf{Bold} denotes the best score for each metric.
    }
    \vspace{-5pt}
    \setlength\tabcolsep{1pt}
    \begin{tabular}{@{}ll cccc c cccc@{}}
    \toprule
    &Dataset & \multicolumn{4}{c}{VITON-HD} && \multicolumn{4}{c}{DressCode} \\
    \cmidrule{3-6} 
    \cmidrule{8-11}
    &Method & LPIPS\,$\downarrow$ & SSIM\,$\uparrow$ & FID\,$\downarrow$ & CLIP-I\,$\uparrow$ && LPIPS\,$\downarrow$ & SSIM\,$\uparrow$ & FID\,$\downarrow$ & CLIP-I\,$\uparrow$ \\
    \midrule
    \rowcolor{RowHighlight}\multicolumn{11}{c}{GAN-based methods} \\
    \midrule
    &HR-VITON~\cite{lee2022high} & 0.115 & 0.883 & 9.70 & 0.832 &
    & 0.112 & 0.910& 21.42 & 0.771 \\
    &GP-VTON~\cite{xie2023gp} & 0.105 & \textbf{0.898} & 6.43 & 0.874 &
    & 0.484 &0.780 &55.21 &0.628 \\
    \midrule
    \rowcolor{RowHighlight}\multicolumn{11}{c}{Diffusion-based methods} \\
    \midrule
    &LaDI-VTON~\cite{morelli2023ladi} 
    & 0.156 &0.872 & 8.85 & 0.834 &
    & 0.149 & 0.905 & 16.54 & 0.803 \\
    &DCI-VTON~\cite{gou2023taming} 
    &   0.166  &0.856 & 8.73 & 0.840 &
    & 0.162 &0.893 & 17.63 &0.777 \\
    &StableVITON~\cite{kim2023stableviton}
    & 0.133  & 0.885 & 6.52 & 0.871 &
    & 0.107 & 0.910 & 14.37 & 0.866 \\
    &\sname~(ours)
    & \textbf{0.102}   & 0.870  & \textbf{6.29} & \textbf{0.883} & 
    & \textbf{0.062}   &  \textbf{0.920}  & \textbf{8.64} &\textbf{0.904}
    \\
    \bottomrule
    \end{tabular}
    \label{tab:main1}
    \vspace{-10pt}
\end{table}

%% file: figures/main/qualwildmain.tex
\begin{figure}[t]
    \small\centering
    \includegraphics[width=0.98\textwidth]{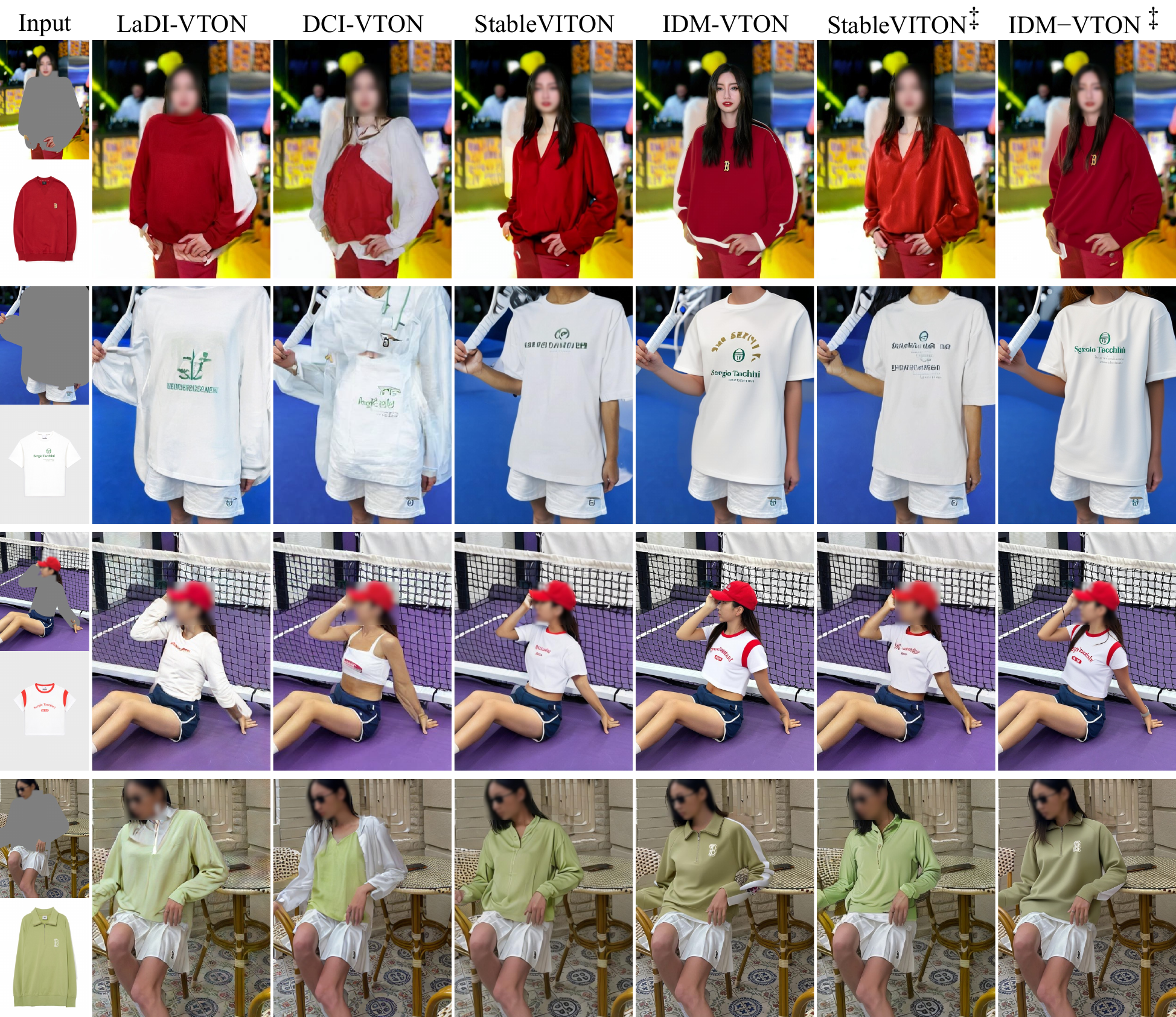}
    \vspace{-5pt}
    \caption{
    \textbf{Qualitative comparisons on In-the-Wild dataset.}
    We show generated virtual try-on images on In-the-Wild dataset using {\sname} (ours) compared with other methods. {\sname} outperforms other methods in generating authentic images and preserving fine-grained details of garment. In particular, customizing {\sname}, (\emph{i.e.}, {\sname}$^\ddag$), significantly enhances the image quality and garment fidelity. When applying customization on StableVITON, (\emph{i.e.}, StableVITON$^\ddag$), the improvements are marginal compared to ours. Best viewed in zoomed, color monitor.
    }
    \label{fig:qual_wildmain}
    \vspace{-15pt}
\end{figure}

%% file: figures/table_wild.tex
\begin{table}[t]
    \centering\small
    \caption{
    \textbf{Quantitative results on In-the-Wild dataset.}
    We compare {\sname} (ours) with other methods on In-the-Wild dataset to assess the generalization capabilities. We report LPIPS, SSIM and CLIP~\cite{radford2021learning} image similarity scores. For {\sname} and StableVITON~\cite{kim2023stableviton}, we further customize models using a single pair of person-garment images (denoted by $^\ddag$). We see that {\sname} outperforms other methods, and customized {\sname} (\emph{i.e.}, {\sname}$^\ddag$) performs the best.
    }
    \vspace{-5pt}
    \begin{tabular}{lccc}
    \toprule
    Method & LPIPS\,$\downarrow$ & SSIM\,$\uparrow$ &  CLIP-I\,$\uparrow$\\
    \midrule
    HR-VITON~\cite{lee2022high} & 0.330 &0.741 & 0.701\\
    LaDI-VTON~\cite{morelli2023ladi} & 0.303 & 0.768 &0.819 \\
    DCI-VTON~\cite{gou2023taming} & 0.283 & 0.735 & 0.752\\
    Stable-VITON~\cite{kim2023stableviton} & 0.260 &0.736 &0.836\\
    {\sname} (ours)& \underline{0.164}  & \underline{0.795} & \underline{0.901} \\
    \midrule
    Stable-VITON$^\ddag$~\cite{kim2023stableviton} & 0.259 &0.733  &0.858\\
    {\sname$^\ddag$} (ours)& \textbf{0.150} & \textbf{0.797} &\textbf{0.909} \\  
    \bottomrule
    \end{tabular}
    \label{tab:table_wild}
\end{table}

%% file: figures/main/ablation_garment.tex
\begin{figure}[t]
    \small\centering
    \includegraphics[width=0.98\textwidth]{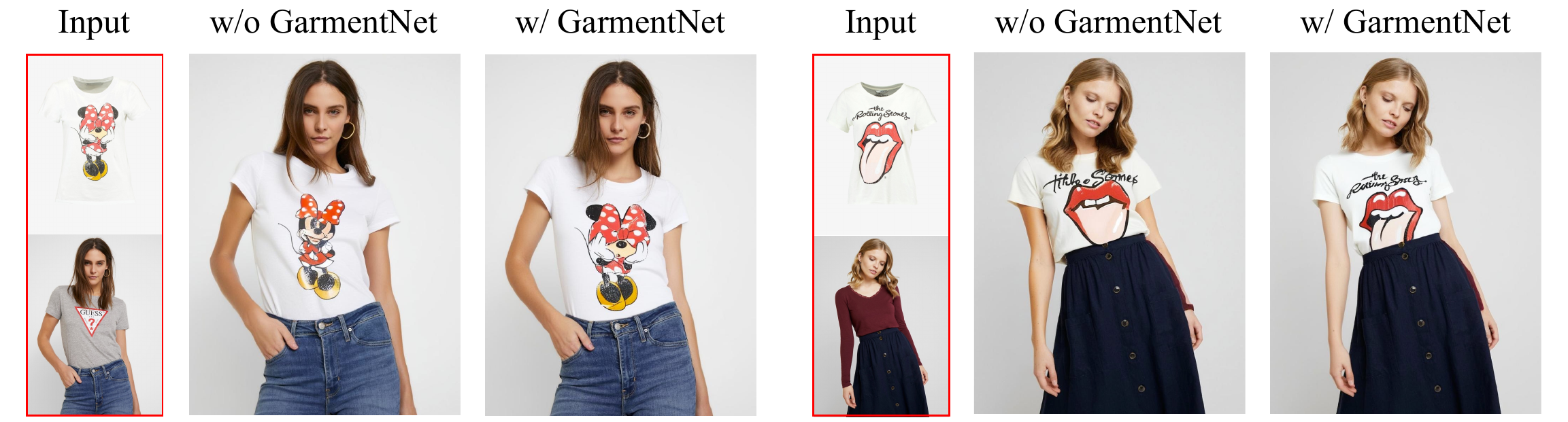}
    \vspace{-5pt}
    \caption{
    \textbf{Effect of GarmentNet.} 
    We compare the generated virtual try-on images without using GarmentNet (left) and together with GarmentNet (right). We observe that using GarmentNet significantly improves retaining the fine-grained details of garment (\emph{e.g.}, the graphics in the t-shirts).
    }
    \label{fig:ablation_garment}
    \vspace{-10pt}
\end{figure}

%% file: figures/main/ablation_caption.tex
\begin{figure}[t]
    \small\centering
    \includegraphics[width=0.98\textwidth]{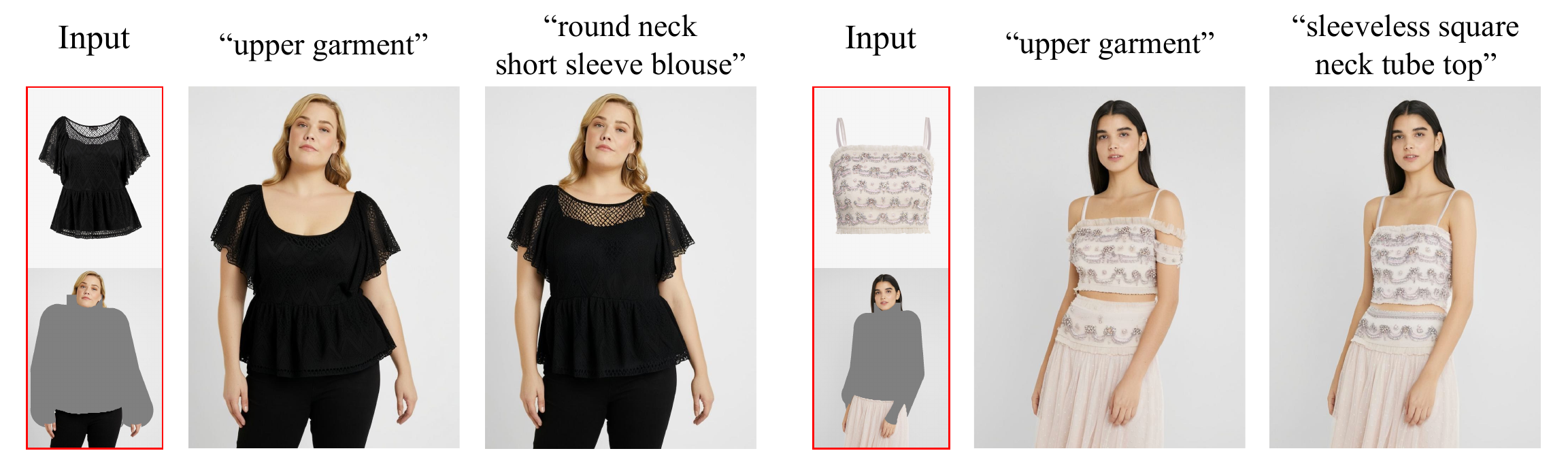}
    \vspace{-5pt}
    \caption{
    \textbf{Effect of detailed caption for garment.} 
    We compare the generated virtual try-on images using na\"ive captions (left) and detailed captions (right). 
    The model trained with detailed captions generates images that are more consistent with the garment. Best viewed in zoomed, color monitor.
    }
    \label{fig:ablation_caption}
    \vspace{-10pt}
\end{figure}

%% file: figures/table_unetclip.tex
\begin{table}[t]
    \centering\small
    \begin{minipage}[t]{0.48\linewidth}
        \centering
        \caption{\textbf{Ablation on GarmentNet.} We ablate the effect of using GarmentNet by comparing LPIPS, SSIM, and CLIP image similarity scores on the VITON-HD test dataset.} 
        \begin{tabular}{cccc}
        \toprule
        GarmentNet & LPIPS\,$\downarrow$  & SSIM\,$\uparrow$ &  CLIP-I\,$\uparrow$\\
        \midrule
        \xmark & 0.121 & 0.849&0.846 \\
        \cmark & \textbf{0.102} & \textbf{0.870} & \textbf{0.883} \\
        \bottomrule
        \label{tab:ablunet}
        \end{tabular}
        
    \end{minipage}
    \hfill
    \begin{minipage}[t]{0.49\linewidth}
        \centering\small
        \caption{
        \textbf{Ablation on customization.}
        We compare default \sname~with different customization methods by fine-tuning all UNet layers, LoRA~\cite{hu2021lora} layers, or up-block layers (Decoder).
        }
        \vspace{-10pt}
        \begin{tabular}{lccc}
        \toprule
        Method & LPIPS\,$\downarrow$ & SSIM\,$\uparrow$ & CLIP-I\,$\uparrow$\\
        \midrule
        Default & 0.1642 &0.795 & 0.901 \\
        \midrule
        All UNet &0.1512 & \textbf{0.797} &0.891 \\
        LoRA & 0.1573 & 0.794 & 0.881\\
        Decoder & \textbf{0.1502} & \textbf{0.797} & \textbf{0.909} \\
        \bottomrule
        \label{tab:adapt}
        \end{tabular}
    \end{minipage}
    \vspace{-10pt}
\end{table}

%% file: appendix.tex
\clearpage
\appendix
\begin{center}
{\bf {\LARGE Appendix}}
\end{center}
 

\section{Implementation Details}
\subsection{In-the-Wild Dataset}
\input{figures/appendix/datasetexp}

In-the-Wild dataset comprises multiple human images wearing each target garment.
Images of garment are collected from MLB online shopping mall\footnote{\href{https://www.mlb-korea.com/main/mall/view}{https://www.mlb-korea.com/main/mall/view}} and images of human wearing each garment are gathered from social media platforms like Instagram.
As shown in Fig.~\ref{fig:dataset_exp}, the human images exhibit diverse backgrounds, ranging from parks and buildings to snowy landscapes. For preprocessing, we employ center cropping on the human images to achieve resolutions of $1024\times768$ resolutions, while the garment images are resized to the same dimensions for compatible setting.

\subsection{Training and Inference}
We train the model using the Adam~\cite{kingma2014adam} optimizer with a fixed learning rate of 1e-5 over 130 epochs (63k iterations with batch size of 24). It takes around 40 hours in training with 4$\times$A800 GPUs. We apply data augmentations following StableVITON~\cite{kim2023stableviton}, where 
we apply horizontal flip~(with probability $0.5$), random affine shifting and scaling~(limit of $0.2$, with probability $0.5$) to the inputs of TryonNet, \emph{i.e.}, $\mathbf{x}_p, \mathbf{x}_{\textrm{pose}}, \mathbf{x}_m$ and $\mathbf{m}$. For customization, we fine-tune our model using the Adam optimizer with a fixed learning rate of 1e-6 for 100 steps. It takes around 2 minutes with a single A800 GPU. 

During the inference, we generate images using the DDPM scheduler with 30 steps. 
We set the strength value to 1.0, \emph{i.e.}, denoising begins from random Gaussian noise, to ignore the masked portion of the input image. For classifier-free guidance~\cite{ho2022classifier} (CFG), we merge both conditioning, \emph{i.e.}, low-level features $\mathbf{L}_g$ from GarmentNet and high-level semantics $\mathbf{H}_g$ from IP-Adapter, as these conditions contain features of the same garment image, following SpaText~\cite{avrahami2023spatext}. In specific, the forward is given as follows: 
\begin{equation}
    \hat{\boldsymbol{\epsilon}}_\theta(\mathbf{x}_t;\mathbf{L}_g,\mathbf{H}_g,t) = s\cdot (\boldsymbol{\epsilon}_\theta(\mathbf{x}_t;\mathbf{L}_g,\mathbf{H}_g,t) - \boldsymbol{\epsilon}_\theta (\mathbf{x}_t;t)) + \boldsymbol{\epsilon}_\theta(\mathbf{x}_t;t)\text{,}
\end{equation}
where $\boldsymbol{\epsilon}_\theta(\mathbf{x}_t;\mathbf{L}_g,\mathbf{H}_g,t)$ denotes noise prediction with conditions $\mathbf{L}_g$ and $\mathbf{H}_g$, and  $\boldsymbol{\epsilon}_\theta(\mathbf{x}_t;t)$ denotes unconditional noise prediction.
We use guidance scale $s$=2.0, which works well in practice.

\subsection{Detailed captioning of garments}
\input{figures/appendix/caption_ex}

We generate detailed captions for each garment to leverage the prior knowledge of T2I diffusion models. We employ OMNIOUS.AI's commercial \footnote{\href{https://omnicommerce.ai}{https://omnicommerce.ai}} fashion attribute tagging annotator, which has been trained with over 1,000 different fashion attributes. The image annotator provides various feature categories present in a given image, such as sleeve length and neckline type. 
We extract three different feature categories: sleeve length, neckline type, and item name, as illustrated in Fig.~\ref{fig:caption_ex}. Subsequently, we generate captions based on this feature information, for example, ``short sleeve off shoulder t-shirts''. 

\input{figures/appendix/ootd_cmp_viton}

\section{Comparison with Concurrent Work}
\input{figures/appendix/ootd_cmp_wild}

\input{figures/table_viton_appendix}

\input{figures/table_dc_appendix}

\input{figures/table_wild_appendix}

We additionally compare {\sname} with OOTDiffusion~\cite{xu2024ootdiffusion}, which is a concurrent work on virtual try-on task. We compare each model trained on VITON-HD and DressCode datasets. We use publicly available model checkpoints for generating try-on images of OOTDiffusion\footnote{\href{https://github.com/levihsu/OOTDiffusion}{https://github.com/levihsu/OOTDiffusion}}.

Fig.~\ref{fig:appendix_ootd_viton} and Fig.~\ref{fig:appendix_ootd_wild} present qualitative comparisons of OOTDiffusion and ours on VITON-HD, DressCode and In-the-Wild dataset.
As shown in Fig.~\ref{fig:appendix_ootd_viton}, we see that {\sname} outperforms OOTDiffusion in capturing both high-level semantics and low-level details, and generating more authentic images.
In particular, we observe notable improvements of {\sname} compared to OOTDiffusion on In-the-Wild dataset, which demonstrates the generalization ability of {\sname}.


Tab.~\ref{tab:appendix_viton}, Tab.~\ref{tab:appendix_dc} and Tab.~\ref{tab:appendix_wild} show quantitative comparisons between {\sname} and OOTDiffusion on VITON-HD, DressCode, and In-the-Wild datasets.
One can notify that {\sname} outperforms OOTDiffusion on all metrics including image fidelity (FID), and reconstruction of garments (LPIPS, SSIM and CLIP-I), which verifies our claim.
\input{figures/appendix/appendix_viton}

\input{figures/appendix/qual_wild_appendix_cmp}

\input{figures/appendix/qualvitonappend}

\input{figures/appendix/qualdc}

\input{figures/appendix/qual_wild1}

\input{figures/appendix/qual_wild2}

\input{figures/appendix/qual_wild3}

\clearpage

%% file: figures/appendix/datasetexp.tex
\begin{figure}[ht]
    \small\centering
    \includegraphics[width=0.98\textwidth]{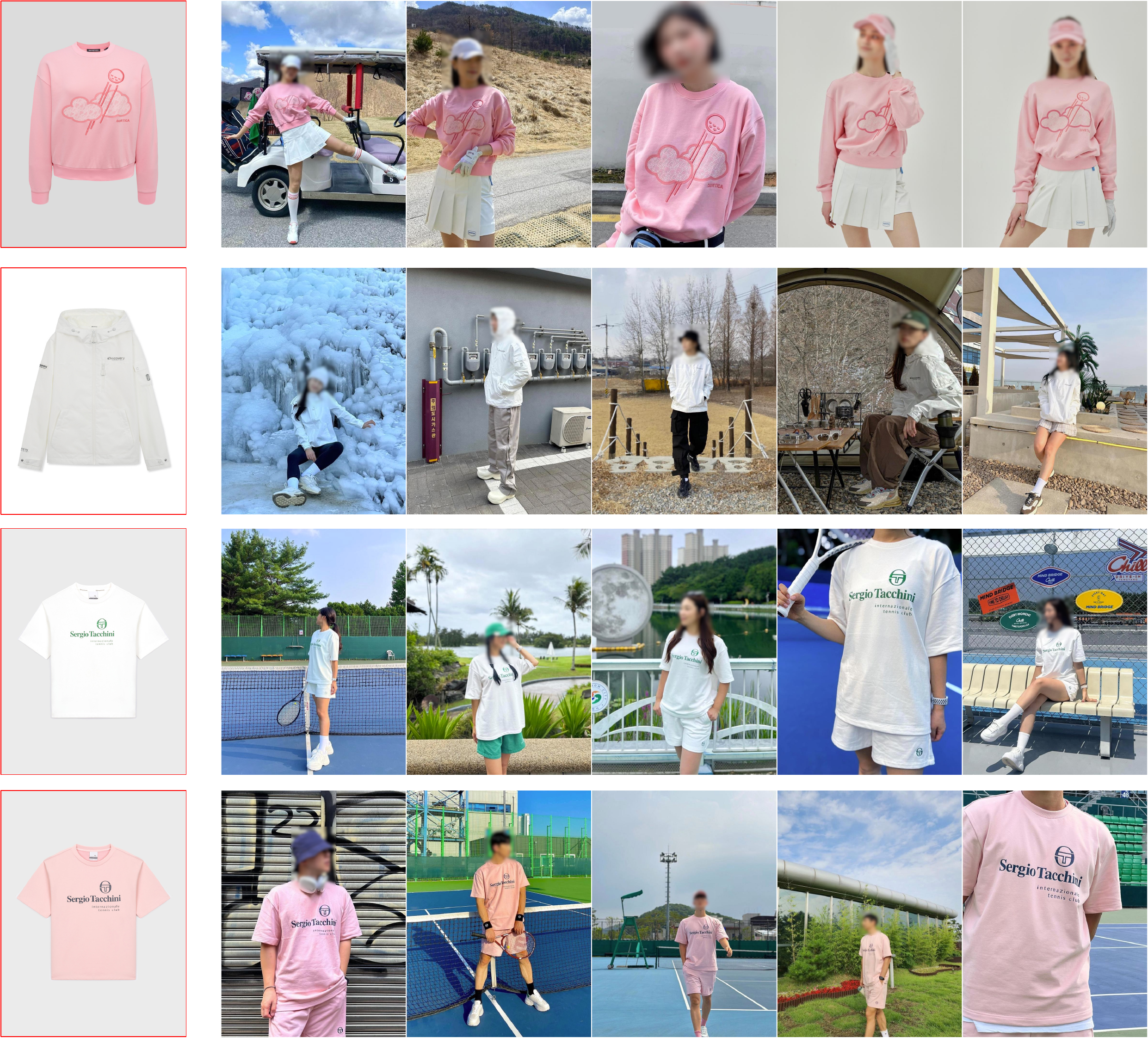}
    \caption{
    Examples of In-the-Wild dataset. We collect pairs of garment and human wearing the garment. 
    } 
    \label{fig:dataset_exp}
\end{figure}

%% file: figures/appendix/caption_ex.tex
\begin{figure}[t]
    \small\centering
    \includegraphics[width=0.98\textwidth]{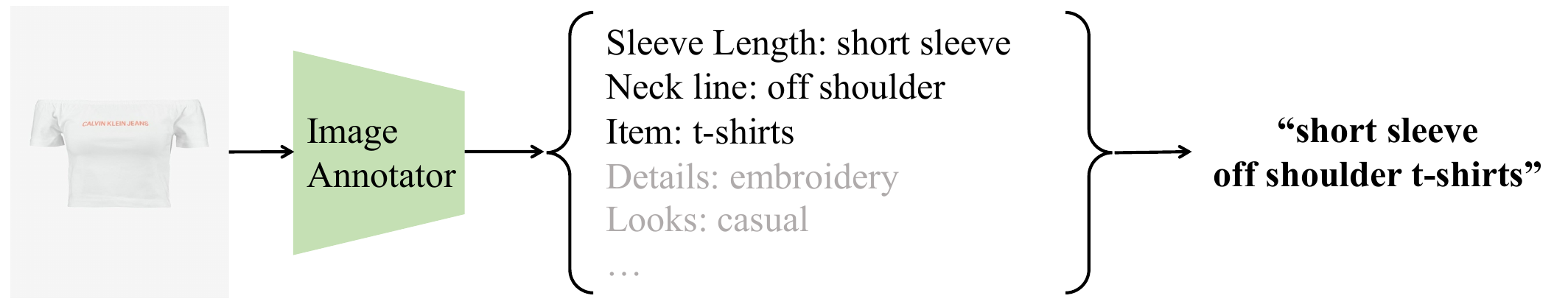}
    \caption{
    \textbf{Illustration of generating detailed captions of garments}. We utilize pretrained fashion attribute tagging annotator to extract information of garment and generate detailed captions of the garment based on extracted information.
    } 
    \label{fig:caption_ex}
\end{figure}

%% file: figures/appendix/ootd_cmp_viton.tex
\begin{figure}[h]
    \small\centering
    \includegraphics[width=0.9\linewidth]{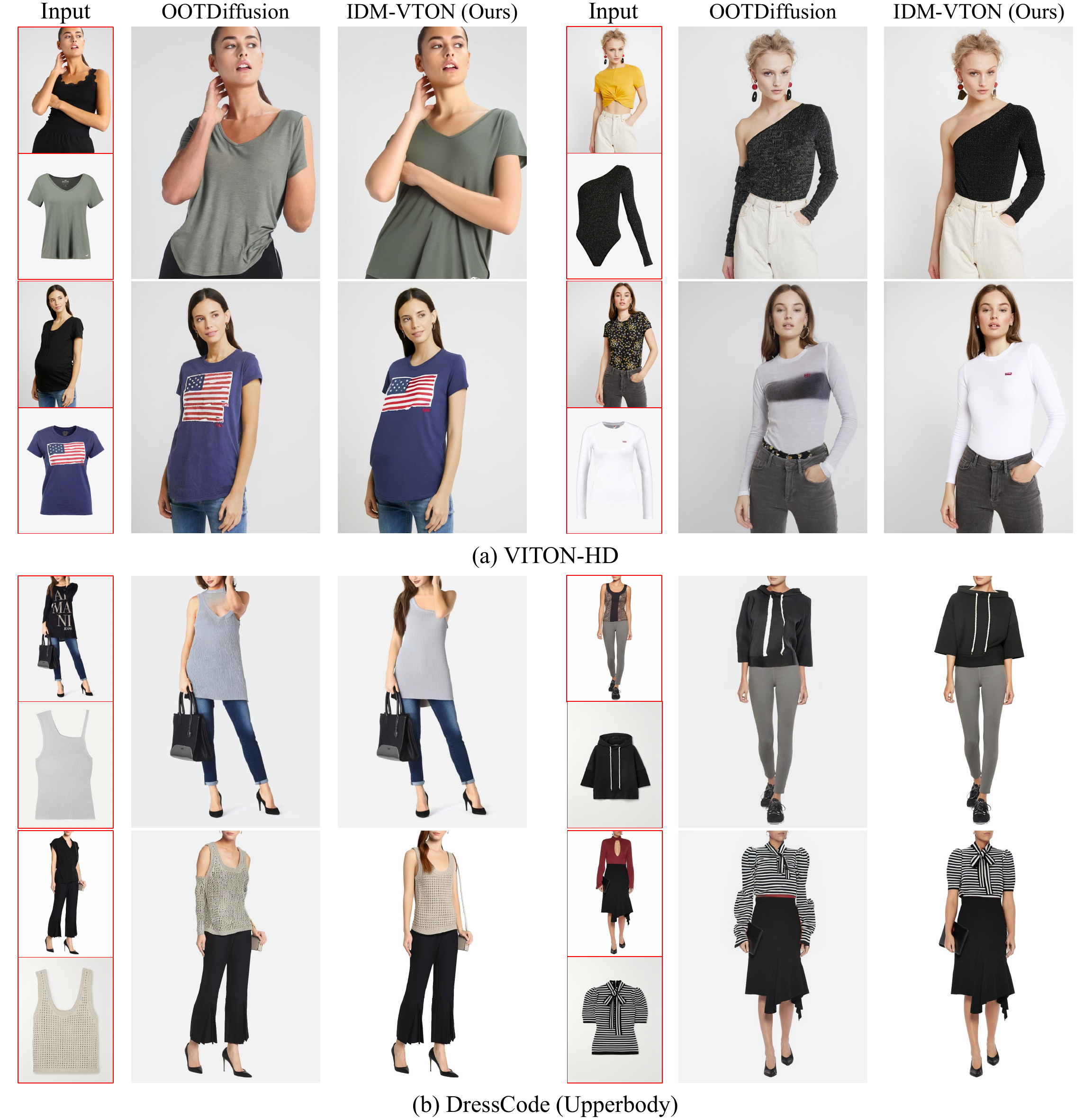}
    \caption{
    \textbf{Comparison between \sname~and OOTDiffusion~\cite{xu2024ootdiffusion} on VITON-HD and DressCode dataset.} 
    Both methods are trained on VITON-HD training dataset.
    Best viewed in zoomed, color monitor.
    }
    \label{fig:appendix_ootd_viton}
    \vspace{-10pt}
\end{figure}

%% file: figures/appendix/ootd_cmp_wild.tex
\begin{figure}[t]
    \small\centering
    \includegraphics[width=0.99\linewidth]{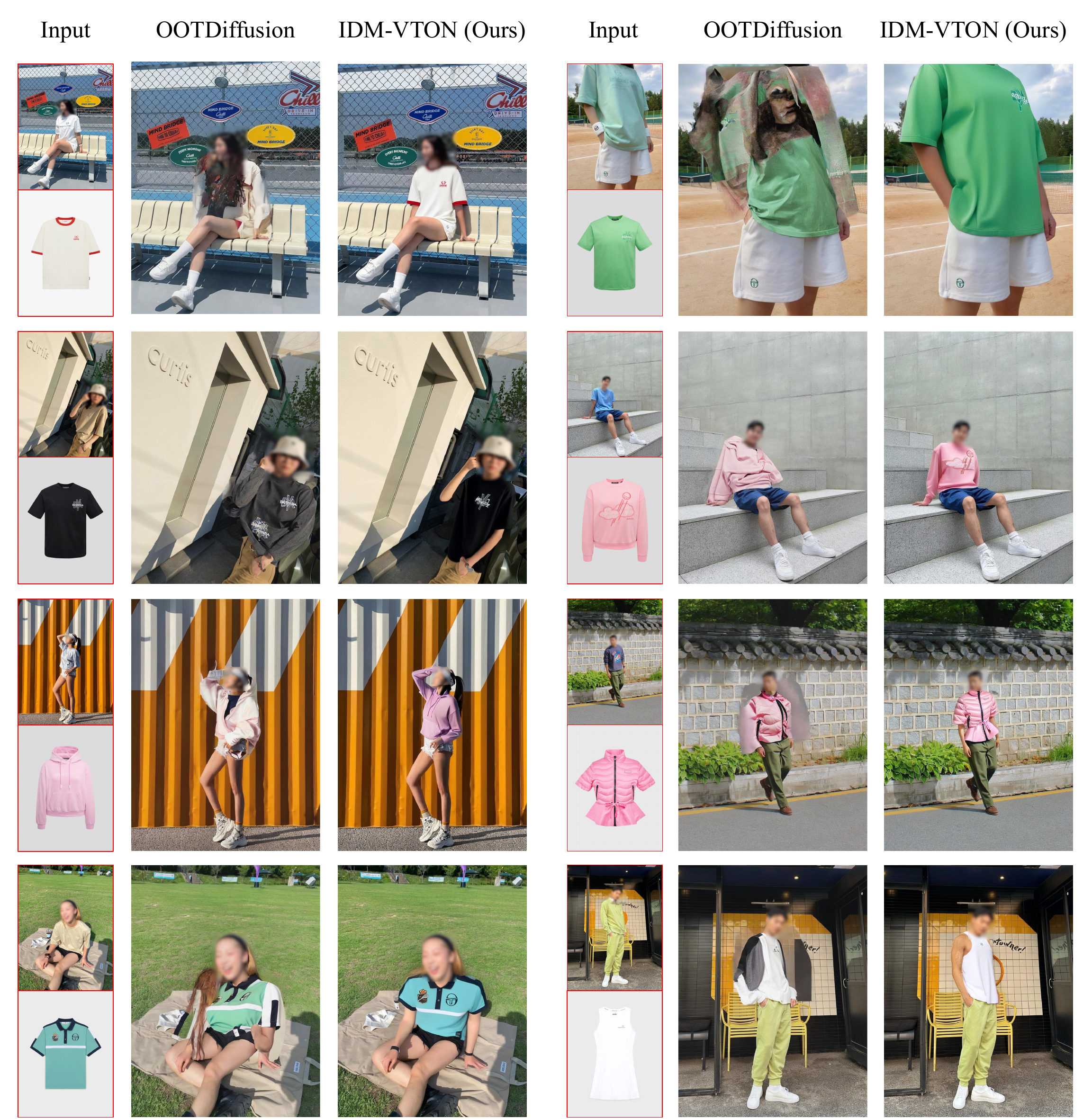}
    \caption{
    \textbf{Comparison between \sname~and OOTDiffusion~\cite{xu2024ootdiffusion} on In-the-Wild dataset.} 
    We provide visual comparison results on In-the-Wild dataset. Both methods are trained on VITON-HD training dataset.
    Best viewed in zoomed, color monitor.
    }
    \label{fig:appendix_ootd_wild}
\end{figure}

%% file: figures/table_viton_appendix.tex
\begin{table}[ht]
    \centering\small
    \caption{\textbf{Quantitative results of models trained on VITON-HD training dataset and evaluated on VITON-HD and DressCode (upper body) test dataset.}
    We additionally compare the metric scores of {\sname} (ours) with the concurrent work OOTDiffusion~\cite{xu2024ootdiffusion}.
    }
    \vspace{-5pt}
    \setlength\tabcolsep{1pt}
    \begin{tabular}{@{}ll cccc c cccc@{}}
    \toprule
    &Train / Test & \multicolumn{4}{c}{VITON-HD / VITON-HD} && \multicolumn{4}{c}{ VITON-HD / DressCode} \\
    \cmidrule{3-6} 
    \cmidrule{8-11}
    &Method & LPIPS\,$\downarrow$ & SSIM\,$\uparrow$ & FID\,$\downarrow$ & CLIP-I\,$\uparrow$ && LPIPS\,$\downarrow$ & SSIM\,$\uparrow$ & FID\,$\downarrow$ & CLIP-I\,$\uparrow$ \\
    \midrule
    &OOTDiffusion~\cite{xu2024ootdiffusion}
    &  0.120  & 0.846 & 6.39 &0.858&
    &  0.087 & 0.901 & 10.68 & 0.852 \\

    &\sname~(ours)
    & \textbf{0.102}   & \textbf{0.870}  & \textbf{6.29} & \textbf{0.883} & 
    & \textbf{0.062}   &  \textbf{0.920}  & \textbf{8.64} &\textbf{0.904}
    \\
    \bottomrule
    \end{tabular}
    \label{tab:appendix_viton}
\end{table}

%% file: figures/table_dc_appendix.tex
\begin{table}[ht]
    \centering\small
    \caption{\textbf{Quantitative results of models trained on DressCode training dataset and evaluated on VITON-HD and DressCode test dataset.}
    We additionally compare the metric scores of {\sname} (ours) with the concurrent work OOTDiffusion~\cite{xu2024ootdiffusion}.
    }
    \vspace{-5pt}
    \setlength\tabcolsep{1pt}
    \begin{tabular}{@{}ll cccc c cccc@{}}
    \toprule
    &Train / Test & \multicolumn{4}{c}{DressCode / DressCode} && \multicolumn{4}{c}{DressCode / VITON-HD} \\
    \cmidrule{3-6} 
    \cmidrule{8-11}
    &Method & LPIPS\,$\downarrow$ & SSIM\,$\uparrow$ & FID\,$\downarrow$ & CLIP-I\,$\uparrow$ && LPIPS\,$\downarrow$ & SSIM\,$\uparrow$ & FID\,$\downarrow$ & CLIP-I\,$\uparrow$ \\
    \midrule
    &OOTDiffusion~\cite{xu2024ootdiffusion}
    &  0.088& 0.885&4.18 &0.851 &
    & 0.205   &  0.788  & 57.87 &0.677 \\
    &\sname~(ours)
    & \textbf{0.072}   &  \textbf{0.902}  & \textbf{3.62} &\textbf{0.913} &
    & \textbf{0.129}   & \textbf{0.859}  & \textbf{12.57} & \textbf{0.839} 
    \\
    \bottomrule
    \end{tabular}
    \label{tab:appendix_dc}
\end{table}

%% file: figures/table_wild_appendix.tex
\begin{table}[ht]
    \centering\small
    \caption{
    \textbf{Quantitative results on In-the-Wild dataset.}
    We compare {\sname} (ours) with the concurrent work OOTDiffusion~\cite{xu2024ootdiffusion} on In-the-Wild dataset to assess the generalization capabilities. We report LPIPS, SSIM and CLIP image similarity scores.
    }
    \vspace{-5pt}
    \begin{tabular}{lccc}
    \toprule
    Method & LPIPS\,$\downarrow$ & SSIM\,$\uparrow$ &  CLIP-I\,$\uparrow$\\
    \midrule
    OOTDiffusion~\cite{xu2024ootdiffusion} &0.203 &0.752 &0.861\\
    {\sname} (ours)& \textbf{0.164}  & \textbf{0.795} & \textbf{0.901} \\
    \bottomrule
    \end{tabular}
    \label{tab:appendix_wild}
\end{table}

%% file: figures/appendix/appendix_viton.tex
\begin{figure}[t]
    \small\centering
    \includegraphics[width=0.99\linewidth]{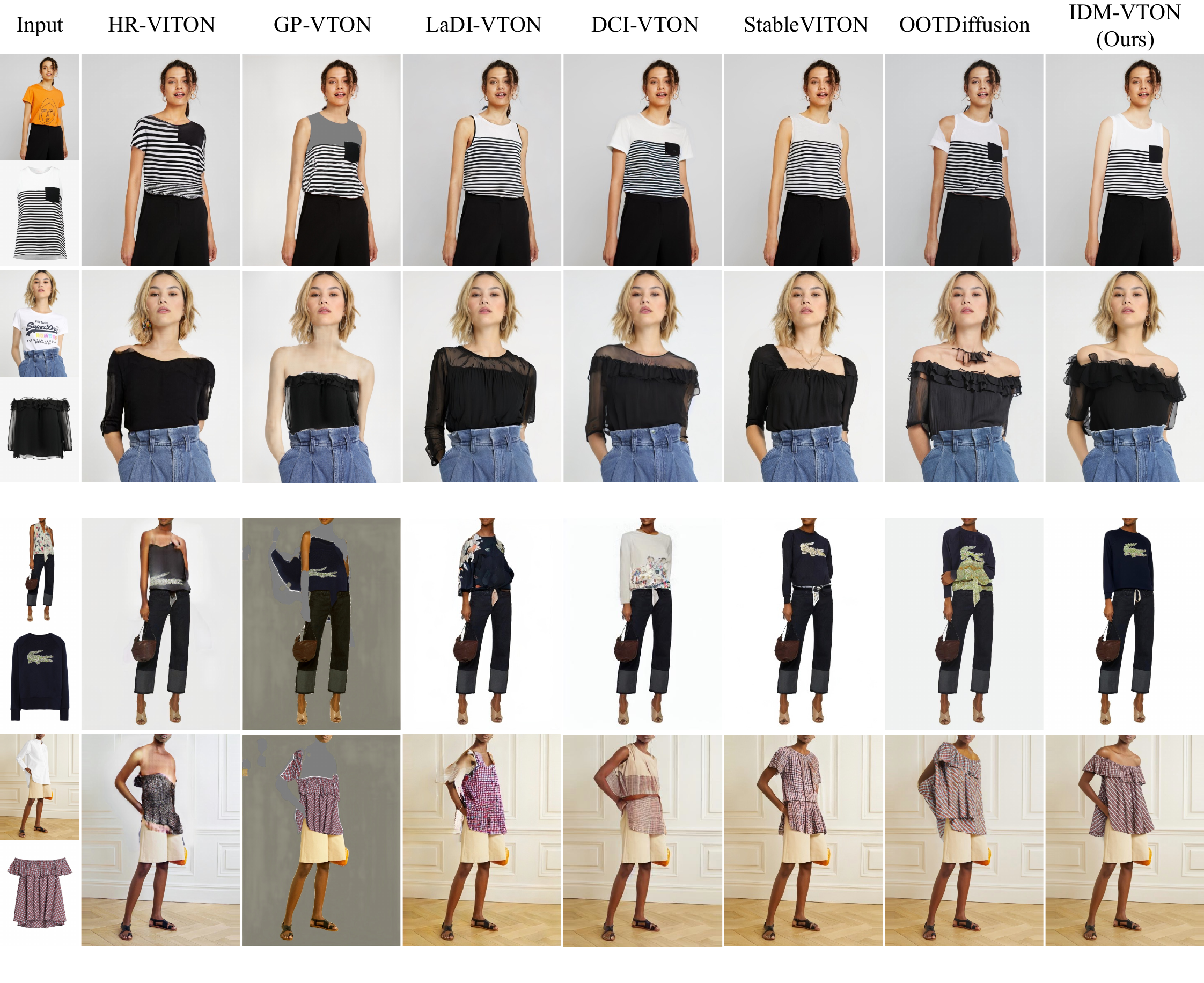}
    \vspace{-5pt}
    \caption{
    \textbf{Qualitative comparison on VITON-HD and DressCode dataset.} 
    As we observed in our quantitative analysis, GAN-based methods generally struggle to generate high-fidelity images introducing non desirable distortions (\emph{e.g.}, non-realistic body and arms) while diffusion-based methods fail to capture low-level features or high-level semantics. All methods are trained on VITON-HD training data.
    }
    \label{fig:appendix_viton}
    \vspace{-10pt}
\end{figure}

%% file: figures/appendix/qual_wild_appendix_cmp.tex
\begin{figure}[t]
    \small\centering
    \includegraphics[width=0.99\textwidth]{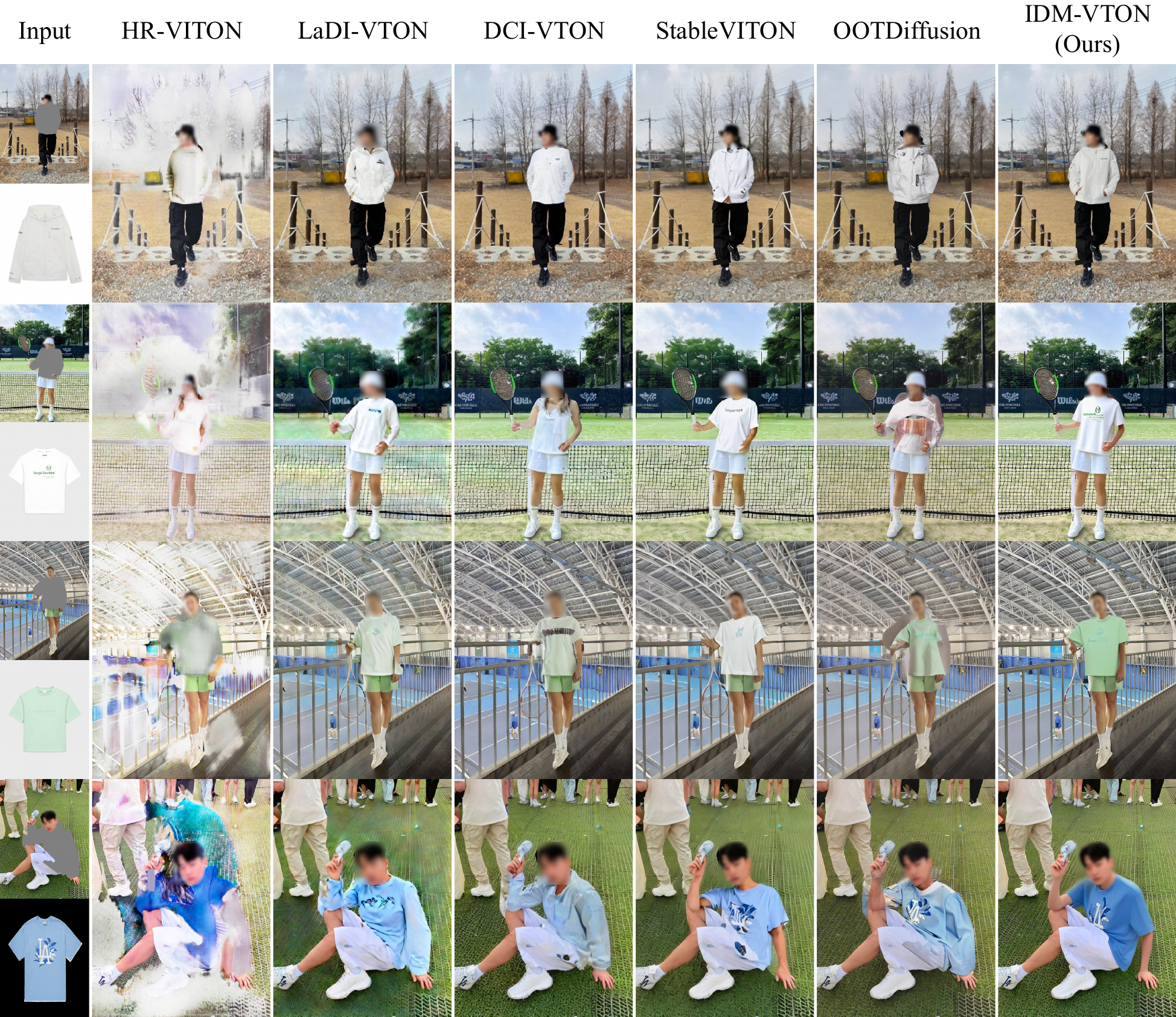}
    \vspace{-5pt}
    \caption{\textbf{Qualitative comparison on In-the-Wild dataset.} While baselines fail to generate natural images or capture the details of clothing, {\sname} produces realistic images and preserves fine details effectively. All methods are trained on VITON-HD training data.
    } 
    \label{fig:qual_wild_appendix_cmp}
    \vspace{-10pt}
\end{figure}

%% file: figures/appendix/qualvitonappend.tex
\begin{figure}[t]
    \small\centering
    \includegraphics[width=0.95\linewidth]{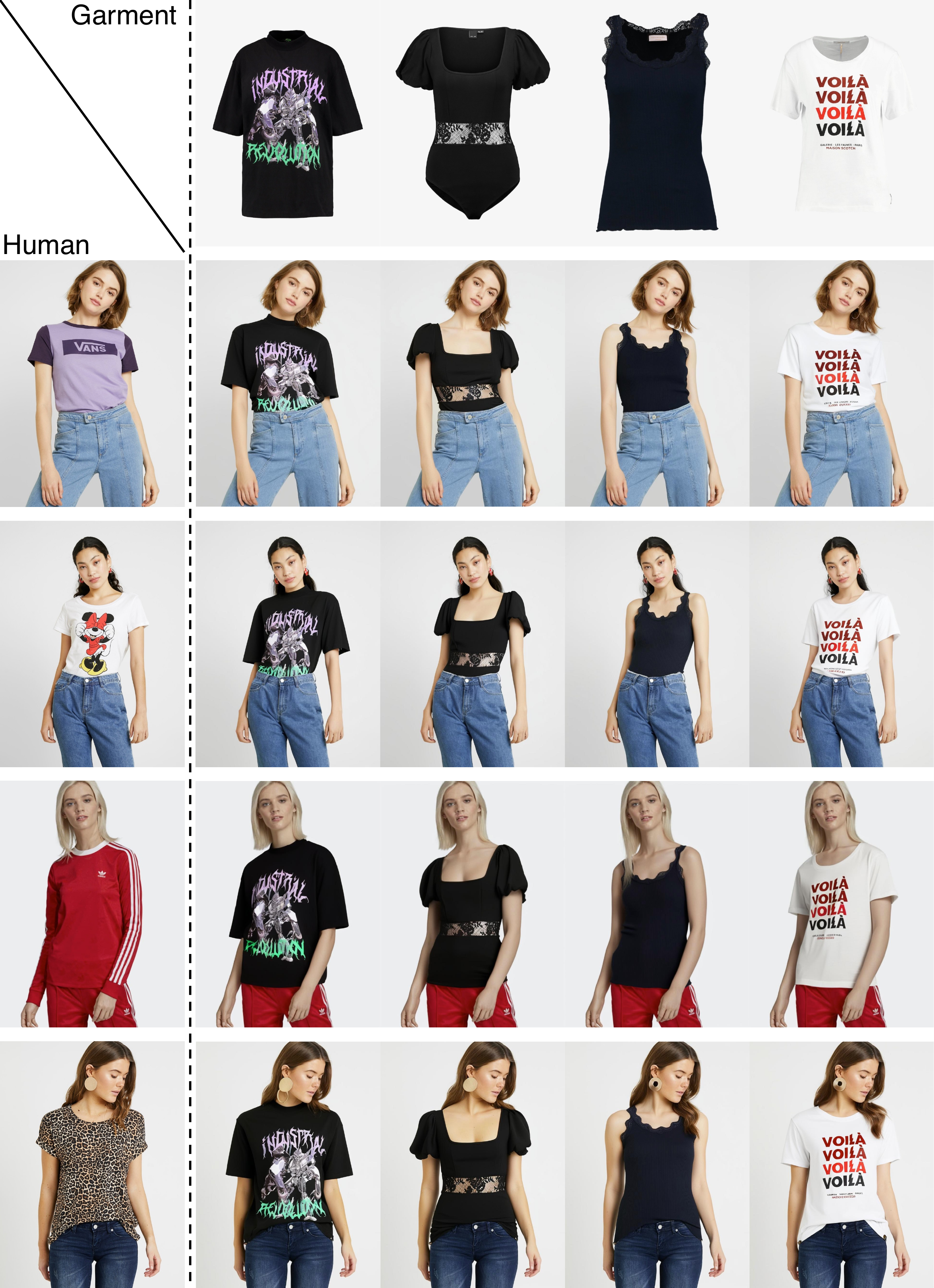}
    \caption{
    Try-on results on VITON-HD test data by \sname~trained on VITON-HD training data. Best viewed in zoomed, color monitor.
    }
    \label{fig:appendix_viton_ours}
\end{figure}

%% file: figures/appendix/qualdc.tex
\begin{figure}[t]
    \small\centering
    \includegraphics[width=0.95\linewidth]{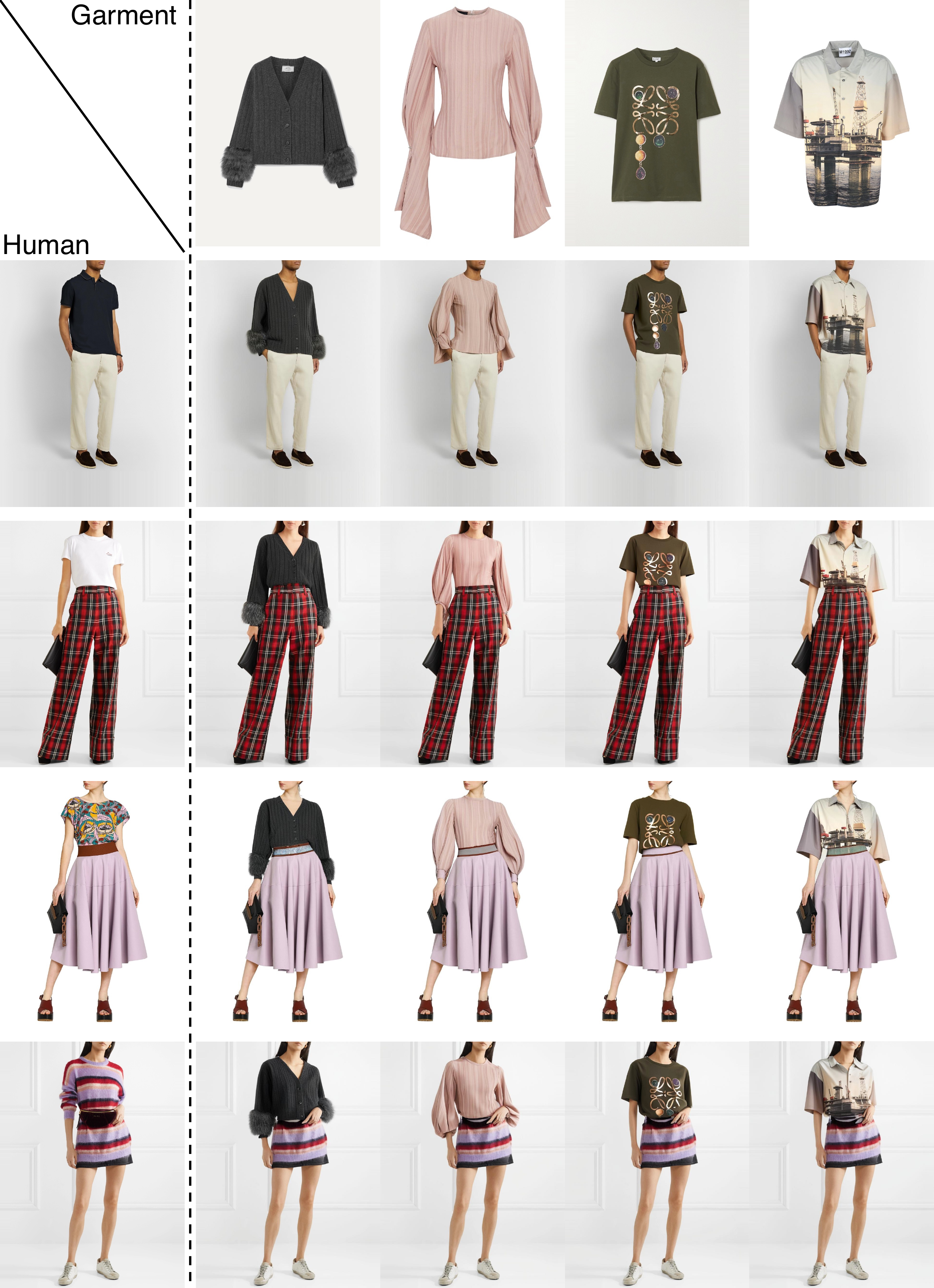}
    \caption{
    Try-on results on DressCode test data by \sname~trained on VITON-HD training data. Best viewed in zoomed, color monitor.
    }
    \label{fig:qualdc}
\end{figure}

%% file: figures/appendix/qual_wild1.tex
\begin{figure}[t]
    \small\centering
    \includegraphics[width=0.95\linewidth]{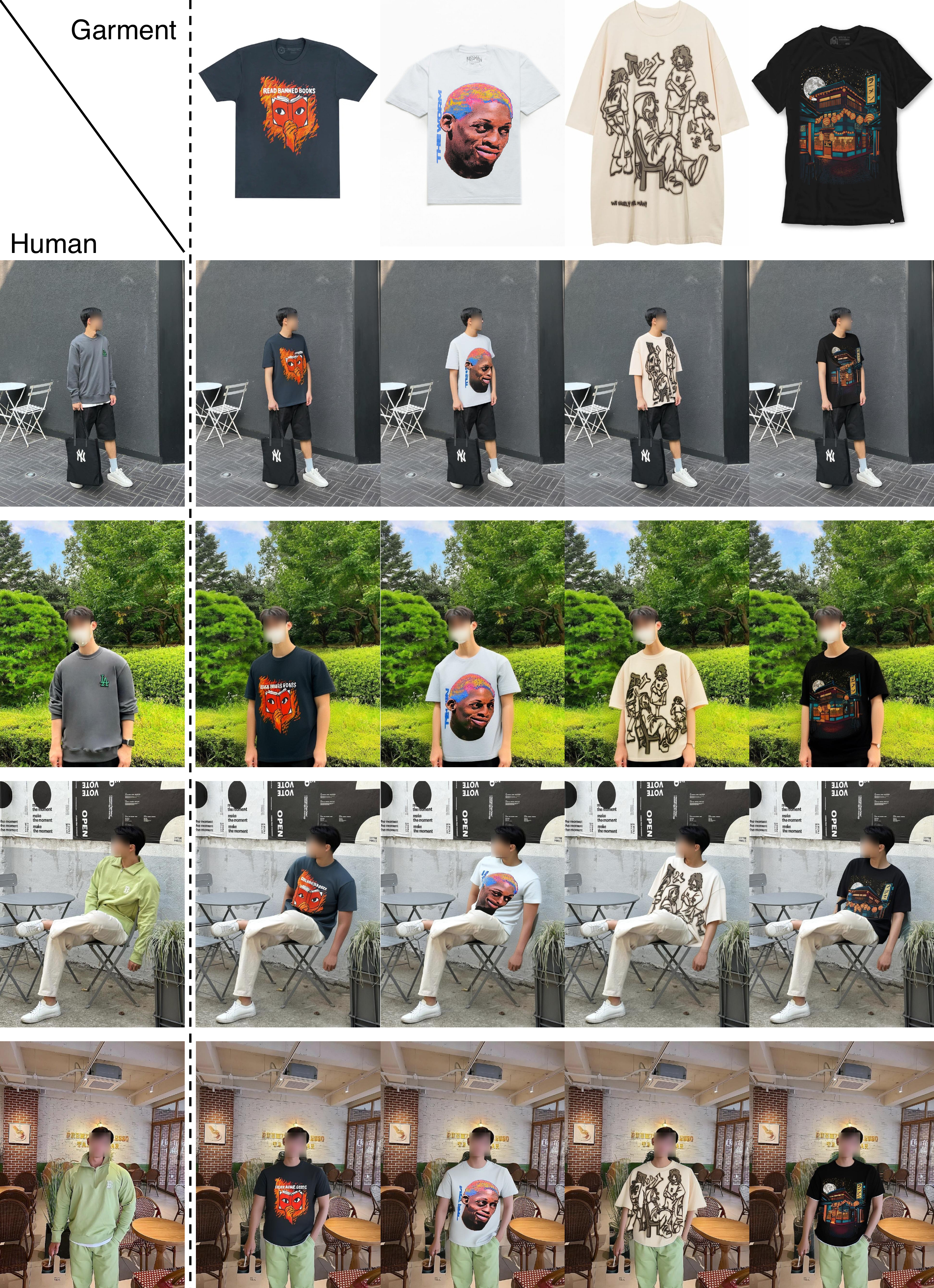}
    \caption{
    Try-on results on In-the-Wild dataset by \sname~trained on VITON-HD training data. Best viewed in zoomed, color monitor.
    }
    \label{fig:appendix_wild_1}
\end{figure}

%% file: figures/appendix/qual_wild2.tex
\begin{figure}[t]
    \small\centering
    \includegraphics[width=0.95\linewidth]{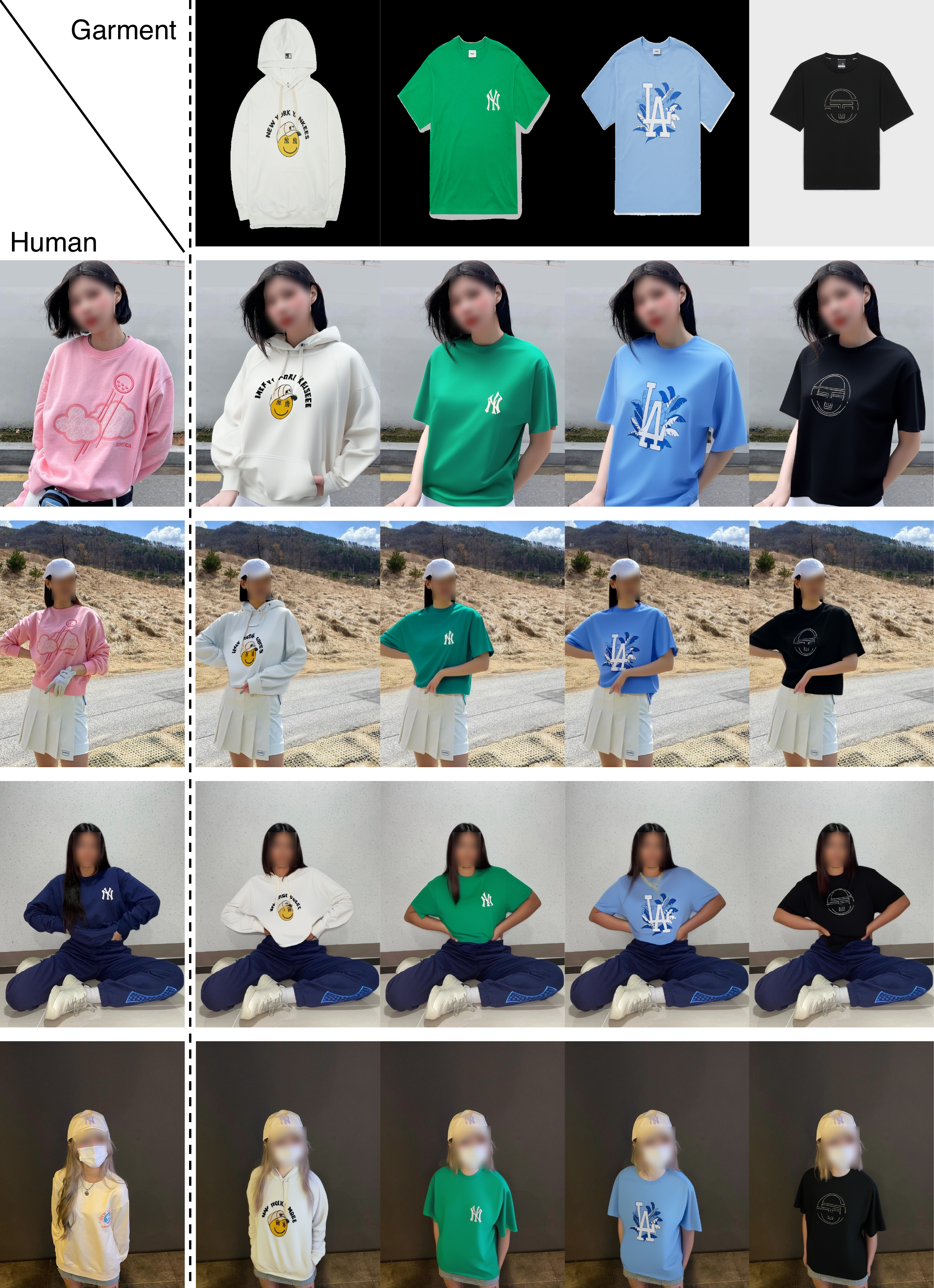}
    \caption{
    Try-on results on In-the-Wild dataset by \sname~trained on VITON-HD training data. Best viewed in zoomed, color monitor.
    }
    \label{fig:appendix_wild_2}
\end{figure}

%% file: figures/appendix/qual_wild3.tex
\begin{figure}[ht]
    \small\centering
    \includegraphics[width=0.95\linewidth]{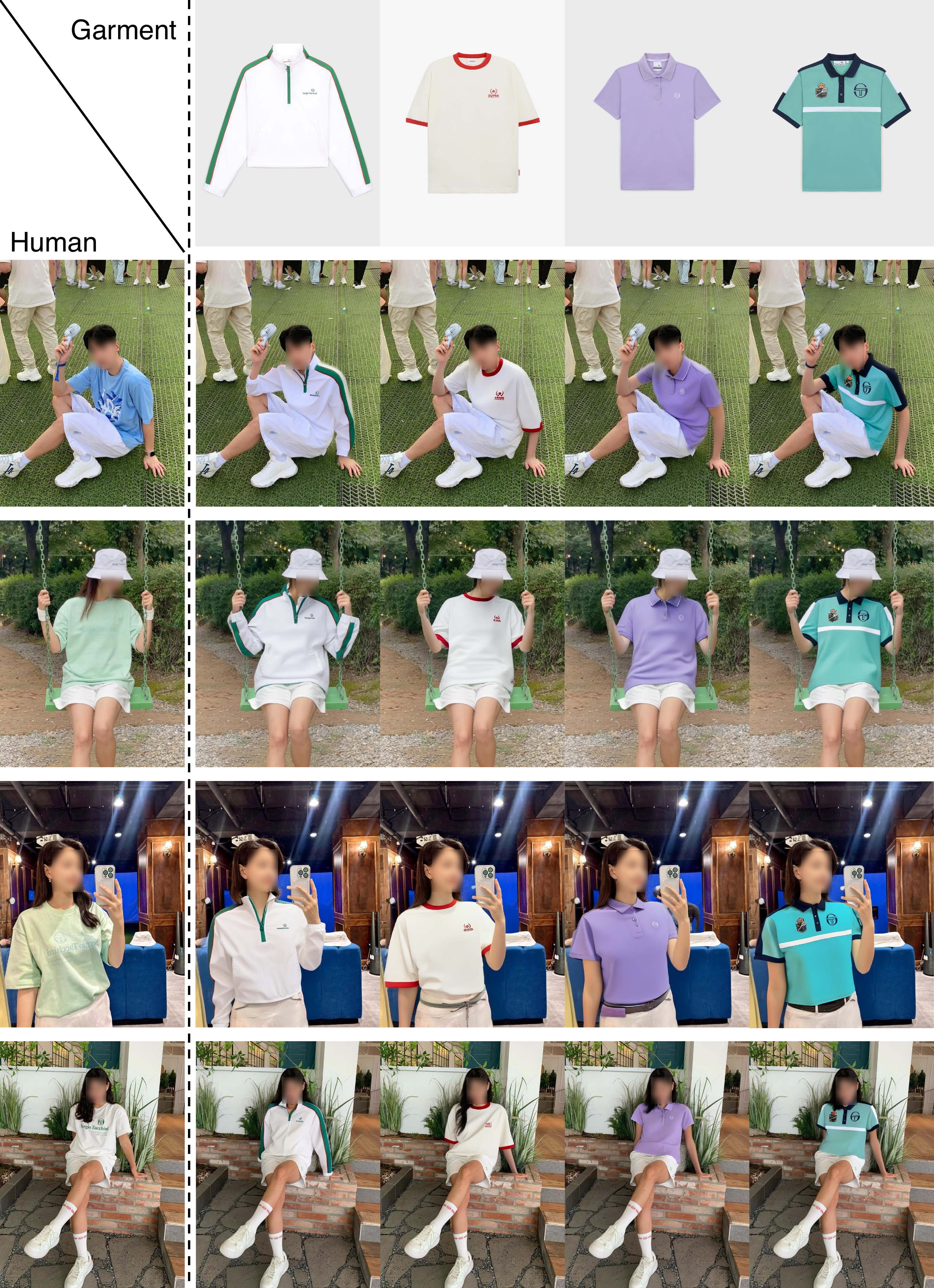}
    \caption{
    Try-on results on In-the-Wild dataset by \sname~trained on VITON-HD training data. Best viewed in zoomed, color monitor.
    }
    \label{fig:appendix_wild_3}
\end{figure}